\documentclass{article} 
\usepackage{iclr2027_conference,times}

\usepackage{amsmath,amsfonts,bm}

\def\eqref#1{equation~\ref{#1}}

\def\1{\bm{1}}

\DeclareMathAlphabet{\mathsfit}{\encodingdefault}{\sfdefault}{m}{sl}
\SetMathAlphabet{\mathsfit}{bold}{\encodingdefault}{\sfdefault}{bx}{n}

\usepackage[utf8]{inputenc}
\usepackage[T1]{fontenc}
\usepackage{newunicodechar}
\newunicodechar{’}{'}
\newunicodechar{‘}{`}
\newunicodechar{“}{``}
\newunicodechar{”}{''}
\usepackage{hyperref}
\usepackage{url}
\usepackage{booktabs}
\usepackage{placeins}
\usepackage{amsfonts}
\usepackage{nicefrac}
\usepackage{microtype}
\usepackage{xcolor}
\usepackage{amsthm}
\usepackage{times}
\usepackage{xspace}
\usepackage{graphicx}
\usepackage{amsmath}
\usepackage{amssymb}
\usepackage{wrapfig}
\usepackage[linesnumbered,ruled,vlined,algo2e]{algorithm2e}
\usepackage{pdfpages}
\usepackage{float} 

\title{VisionLogic: Discovering and Grounding\\Decision-Relevant Visual Concepts}

\author{\hspace*{-\tabcolsep}
\begin{minipage}[t]{\textwidth}
\normalfont\centering
Chuqin Geng$^{1,2}$ \quad
Yuhe Jiang$^{1}$ \quad
Li Zhang$^{1}$ \quad
Zhaoyue Wang$^{2}$ \\[0.2em]
Haolin Ye$^{3}$ \quad
Mark Zhang$^{1}$ \quad
Jingkai Xu$^{4}$ \quad
Xujie Si$^{1}$ \\[0.5em]
$^{1}$University of Toronto \\
$^{2,3}$McGill University \\
$^{4}$McMaster University \\[0.3em]
\texttt{chuqin.geng@mail.mcgill.ca} \quad
\texttt{six@cs.toronto.edu}
\end{minipage}\hspace*{-\tabcolsep}
}

\iclrfinalcopy
\hypersetup{
  hidelinks,
  pdftitle={VisionLogic: Discovering and Grounding Decision-Relevant Visual Concepts},
  pdfauthor={Chuqin Geng, Yuhe Jiang, Li Zhang, Zhaoyue Wang, Haolin Ye, Mark Zhang, Jingkai Xu, Xujie Si}
}
\begin{document}

\maketitle
\lhead{Preprint}

\begin{abstract}
Concept-based explanations help users understand vision models through recognizable visual patterns. However, existing methods often rely on correlational signals without directly validating which image cues support prediction-relevant internal features. To this end, we introduce \textsc{VisionLogic}, a post-hoc framework that grounds these features in visual concepts through intervention-based validation. \textsc{VisionLogic} first identifies compact sets of features whose contributions reproduce the model's original prediction. It then represents their activation states as predicates using class-specific thresholds. An iterative refinement procedure grounds these predicates in visual regions through ablation tests. A region is accepted when its removal deactivates the corresponding predicate, linking the feature's numerical role to visual evidence in the input. The same predicates allow us to examine how features are activated, selected, and reused across images and classes. Across CNNs and vision transformers on ImageNet-1k, we find that only a few features are selected to explain each prediction, and frequently active features are not always selected. In a large-scale human evaluation with 465 participants, \textsc{VisionLogic} significantly improves participants' understanding of model behavior over established methods \textsc{ACE} and \textsc{CRAFT}.
Code is available at \url{https://github.com/allengeng123/VisionLogic}.
\end{abstract}

\section{Introduction}
\label{sec:intro}

Deep learning-based vision models have achieved remarkable success across numerous tasks, yet their \emph{black-box} nature remains a major obstacle to trustworthy AI. This challenge persists as vision architectures evolve from Convolutional Neural Networks (CNNs)~\citep{lecun1998gradient,he2016deep} to Vision Transformers (ViTs)~\citep{dosovitskiy2021image}. To address this issue, a wide range of interpretability methods have been proposed. Among them, concept-based explanations~\citep{nguyen2016synthesizing,BauZKO017,TCAV,ace,craft,CRP} have attracted particular interest because they uncover high-level semantic concepts rather than only providing low-level attribution maps such as Grad-CAM~\citep{selvaraju2017gradcam}.

However, concept discovery often relies on \emph{correlational evidence}, leaving the connection between recognizable concepts and prediction-relevant internal features insufficiently grounded~\citep{Causalsignals,Causalreasoningcv,CaCE}. For instance, \textsc{TCAV}~\citep{TCAV} learns concept directions in activation space and measures prediction sensitivity, while \textsc{ACE}~\citep{ace} applies clustering to identify recurring concepts from image segments. These methods identify meaningful patterns and assess their relevance, but a concept's association with a class does not by itself establish which image cues support a specific internal feature. Without this validation, visually plausible explanations may attribute a prediction to the wrong image cues. This can mislead users about the model's behavior, particularly when identifying biases or understanding its failures.

We address this gap with \textsc{VisionLogic}, a post-hoc framework for discovering and grounding decision-relevant visual concepts through intervention-based validation. Our approach operates in three stages. First, we identify compact sets of features whose contributions through the final classification head reproduce the model's original prediction. Second, we represent their activation states as predicates by fitting class-specific thresholds. This provides an intermediate symbolic representation for understanding how vision models make decisions.

Third, we ground these predicates in visual regions using an iterative procedure that combines neuron-specific attribution maps with ablation tests. By testing whether removing a region deactivates the corresponding predicate, we connect internal features to supporting visual evidence. Foreground segmentation~\citep{zheng2024birefnet} highlights the visual content of these regions, and recurring patterns across images help interpret the features as human-understandable concepts. Together, the three stages provide a common representation for visual explanations and global analysis of feature use.

Our large-scale human evaluation with 465 participants demonstrates the practical value of the discovered concepts. Compared with established concept-based methods \textsc{ACE} and \textsc{CRAFT}, \textsc{VisionLogic} significantly improves participants' understanding of model behavior in tasks involving bias detection and the discovery of non-obvious strategies. Experiments on CNNs and ViTs on ImageNet-1k~\citep{deng2009imagenet} show that features can be frequently active yet rarely selected to explain a prediction. Typically, only a few features are selected for each image, and some features are reused across hundreds of classes. Our contributions are summarized as follows:

\begin{itemize}
    \item We propose \textsc{VisionLogic}, a post-hoc framework that identifies compact sets of features whose contributions reproduce the model's prediction and represents their activation states as predicates for concept-based explanation.

    \item We develop an iterative grounding procedure that combines neuron-specific attribution maps with ablation tests to ground these predicates in visual regions.

    \item We conduct a large-scale human evaluation with 465 participants, demonstrating significant improvements over \textsc{ACE} and \textsc{CRAFT} in participants' understanding of model behavior.

    \item We analyze how features are activated, selected, and reused across CNNs and ViTs, connecting explanations of individual predictions to broader patterns of feature use across classes.
\end{itemize}

\section{Related Work}
\label{sec:back}

Existing approaches to interpreting vision models increasingly focus on \emph{concept-based methods}, which aim to link internal representations to human-interpretable concepts rather than pixel-level attributions~\citep{selvaraju2017gradcam,chattopadhay2018gradcam++,wang2020scorecam}. Early studies explore the \emph{hidden semantics} of neural networks by visualizing what individual neurons or layers encode. For instance, \citet{nguyen2016synthesizing} generate synthetic images that maximally activate specific neurons, while \citet{mahendran2015understanding} reconstruct inputs from intermediate feature maps to reveal the information preserved at different network depths. Network Dissection~\citep{BauZKO017} further provides a systematic framework to quantify the alignment between hidden units and semantic concepts across diverse architectures. These methods offer valuable insights into model representations, motivating further study of how the identified features support individual predictions. Subsequent work establishes explicit links between model activations and human-interpretable concepts. \textsc{Net2Vec}~\citep{fong2018net2vec} trains linear predictors on activation patterns to align feature maps with semantic labels, while \textsc{TCAV}~\citep{TCAV} learns concept directions in activation space and measures prediction sensitivity along these directions. These approaches help identify what a model represents and assess the relevance of predefined concepts.

More recent methods automate concept discovery. For instance, \textsc{ACE}~\citep{ace} applies unsupervised clustering to representations of image segments and scores the resulting concepts using \textsc{TCAV}. It also evaluates concept importance through addition and removal experiments. \textsc{ICE}~\citep{DBLP:conf/aaai/ZhangM0ER21} introduces invertible concept-based explanations using non-negative concept activation vectors to improve interpretability and fidelity. \textsc{CRAFT}~\citep{craft} further combines recursive concept discovery with sensitivity analysis to measure how concept perturbations affect model predictions. Related work on causal concept effects, such as \textsc{CaCE}~\citep{CaCE}, estimates how intervening on concept presence changes classifier predictions.

However, concept discovery often relies on \emph{correlational evidence}, leaving the connection between recognizable concepts and prediction-relevant internal features insufficiently grounded. We address this gap with \textsc{VisionLogic}, which directly tests whether candidate visual regions support the internal features selected to explain a prediction. The resulting explanations help users better understand the visual evidence behind a model's behavior than those provided by \textsc{ACE} and \textsc{CRAFT}, as demonstrated by our large-scale human evaluation.
\section{The \textsc{VisionLogic} Framework}
\label{sec:method}

\textsc{VisionLogic} explains a frozen vision classifier in three stages. First, we identify a compact set of features whose contributions reproduce the model's prediction. Second, we represent their activation states as predicates using class-specific thresholds, allowing comparison across images and classes. Third, we ground these predicates in visual regions through ablation tests.

\subsection{Selecting Compact Sets of Features}
\label{sec:pathways}

Let $z(x)\in\mathbb{R}^{d}$ denote the feature activations supplied to the model's final linear classification head. The logit for class $k$ and the model prediction are
\begin{equation}
 F_k(x)=b_k+\sum_{j=1}^{d}W_{kj}z_j(x),
 \qquad \hat y(x)=\arg\max_{k\in\{1,\ldots,C\}}F_k(x).
 \label{eq:head}
\end{equation}
For an image predicted as $c=\hat y(x)$, the term $W_{cj}z_j(x)$ gives feature $j$'s contribution to the class-$c$ logit. We sort features by decreasing contribution, obtaining the order $\pi_x=(\pi_1,\ldots,\pi_d)$. Starting from the full bias vector, we add their contributions to all class logits in this order:
\begin{equation}
 F_k^{(m)}(x)=b_k+\sum_{r=1}^{m}W_{k\pi_r}z_{\pi_r}(x),
 \qquad k\in\{1,\ldots,C\}.
 \label{eq:partial}
\end{equation}
We stop at the first nonempty prefix that reproduces the original prediction:
\begin{equation}
 m^*(x)=\min\{m\in\{1,\ldots,d\}:\arg\max_k F_k^{(m)}(x)=c\},
 \qquad P(x)=(\pi_1,\ldots,\pi_{m^*(x)}).
 \label{eq:pathway}
\end{equation}
Here, $P(x)$ lists the selected features in contribution order.

The selection procedure accounts for each feature's contribution to both the predicted class and its competitors. In particular, the margin against class $k$ after selecting the first $m$ features is
\begin{equation}
 M_{c,k}^{(m)}(x)=b_c-b_k+
 \sum_{r=1}^{m}(W_{c\pi_r}-W_{k\pi_r})z_{\pi_r}(x).
 \label{eq:margin}
\end{equation}
Keeping the activations of the features in $P(x)$ and setting all others to zero preserves the head's prediction.

We record each selected feature as $(j,+)$ if $z_j(x)\geq0$ and $(j,-)$ otherwise. These pairs form the ordered sequence $S(x)$, preserving the order in $P(x)$. The sign refers to the activation $z_j(x)$ supplied to the head, rather than the contribution $W_{cj}z_j(x)$, and does not assume a particular activation function. Feature selection requires no classifier retraining, gradients, or concept annotations; sorting costs $O(d\log d)$ and accumulating the contributions takes at most $O(Cd)$ operations.

\subsection{Representing Feature Activations as Predicates}
\label{sec:predicates}

The selected features explain individual predictions through their numerical contributions. To compare their activation states across images and ground these features in image cues, we define predicates using class-specific thresholds. We write $p_{c,(j,s)}$ for the predicate associated with class $c$ and signed feature $(j,s)$. When $c$ is clear from context, we abbreviate it as $p_{(j,s)}$. Across classes, identical abbreviated labels refer to the same signed feature but may use different thresholds.

Let $\mathcal D_c^{\mathrm{src}}$ denote the retained source examples assigned to model-predicted class $c$. We use the features selected by the procedure in Section~\ref{sec:pathways}, recorded in $S(x)$, to define the predicates and fit their thresholds. The same ordered sequence of signed features may occur in different classes. These selections explain the model's original predictions and do not assign classes independently.

\paragraph{Class-specific thresholds.}
For each signed feature, we collect the source images in class $c$ for which it is selected:
\begin{equation}
 \mathcal I_{c,(j,s)}=\{x\in\mathcal D_c^{\mathrm{src}}:(j,s)\in S(x)\},
 \qquad \mathcal V_c=\{(j,s):\mathcal I_{c,(j,s)}\ne\varnothing\}.
 \label{eq:predicate-support}
\end{equation}
Thus, $\mathcal V_c$ contains the signed features selected in at least one source image of class $c$. We define a predicate for each of these features, with threshold and Boolean value
\begin{align}
 T_{c,(j,+)}&=\min_{x\in\mathcal I_{c,(j,+)}}z_j(x),
 &p_{c,(j,+)}(x)&=\mathbf{1}\{z_j(x)\geq T_{c,(j,+)}\},\nonumber\\
 T_{c,(j,-)}&=\max_{x\in\mathcal I_{c,(j,-)}}z_j(x),
 &p_{c,(j,-)}(x)&=\mathbf{1}\{z_j(x)\leq T_{c,(j,-)}\}.
 \label{eq:thresholds}
\end{align}
Class-specific thresholds allow the same signed feature to have different activation criteria across classes. This design is motivated by feature polysemanticity: a shared feature may correspond to different visual concepts in different classes.

\paragraph{Active and selected features.}
For an image $x$, we use the predicates defined for its predicted class $c=\hat y(x)$. We distinguish three sets of signed features:
\begin{align}
 A_c(x)&=\{(j,s)\in\mathcal V_c:p_{c,(j,s)}(x)=1\},\nonumber\\
 Q_c(x)&=\operatorname{set}(S(x))\cap\mathcal V_c,\qquad
 J_c(x)=A_c(x)\cap Q_c(x).
 \label{eq:predicate-readout}
\end{align}
Here, $A_c(x)$ contains features whose predicates are active, $Q_c(x)$ contains selected features with predicates defined from the source images, and $J_c(x)$ contains features satisfying both conditions. Selected features outside $\mathcal V_c$ remain part of the numerical explanation but have no fitted predicate for class $c$.

On source images, selection implies activation because the thresholds are fitted from the selected occurrences. For held-out images, we apply the same feature selection procedure, but a selected feature need not satisfy its fixed source threshold. For visual explanation, we ground the predicates associated with $J_c(x)$: their features are selected for the image, and their thresholds are satisfied.

\subsection{Grounding Predicates in Visual Concepts}
\label{sec:grounding}

\begin{figure*}[!t]
    \centering
    \includegraphics[width=\linewidth]{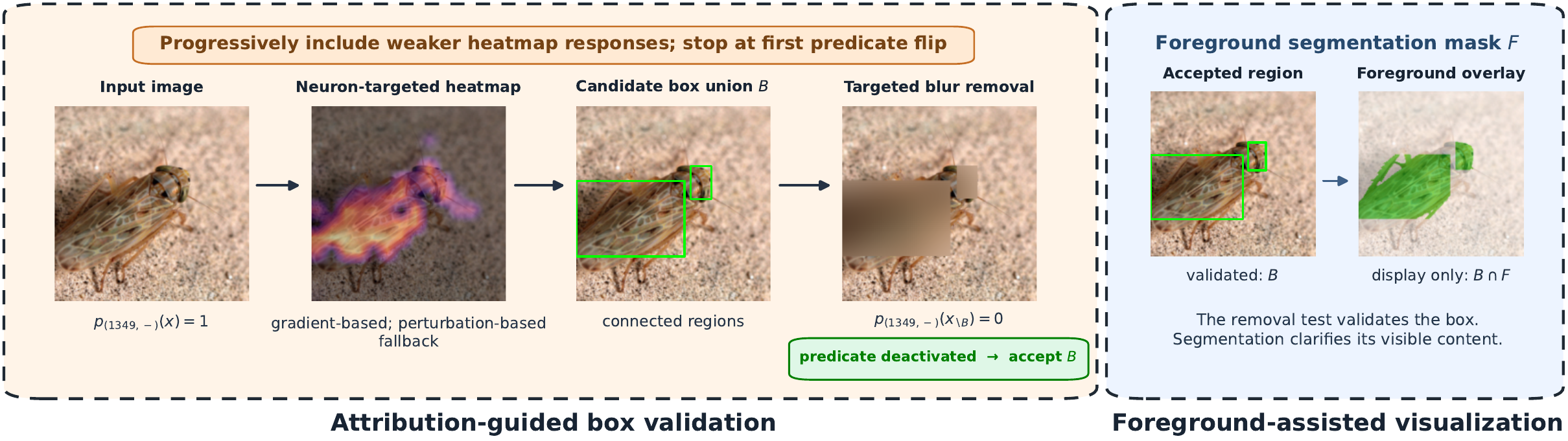}
    \caption{\textbf{Intervention-based visual grounding.}
    Neuron-specific attribution maps guide the search for candidate regions. We ablate these regions by replacing their content with a blurred version. If the predicate remains active, we iteratively refine the proposals by including weaker heatmap responses. The first union of boxes whose ablation deactivates the target predicate is accepted. Foreground segmentation highlights the visual content within the accepted boxes.}
    \label{fig:grounding}
\end{figure*}

The final stage grounds abstract predicates in visual concepts through ablation tests, as illustrated in Figure~\ref{fig:grounding}. Given an image $x$, we consider the predicates associated with $J_c(x)$, which are active and whose features are selected to explain the prediction (Section~\ref{sec:predicates}). For each predicate, we blur a candidate region~\citep{fong2017interpretable} to obtain $x'$ and recompute the predicate. A change from active to inactive provides evidence that the region supports its activation, directly connecting the selected internal feature to visual evidence in the image.

To locate these regions, we use neuron-specific attribution maps that target the predicate's signed neuron activation rather than the final class score. We begin with Grad-CAM~\citep{selvaraju2017gradcam} and draw tight bounding boxes around connected areas with strong heatmap responses. When several disjoint areas are highlighted, their boxes are ablated together. If the ablation does not deactivate the predicate, we progressively include areas with weaker heatmap responses. If the Grad-CAM search fails, we repeat the search using another attribution method such as Score-CAM~\citep{wang2020scorecam}. Using different attribution methods provides complementary proposals and can expand the search space. We accept the first region whose ablation deactivates the predicate. If all searches fail, the predicate remains unlocalized for that image. In short, attribution guides where to search, while the ablation test determines which regions are accepted.

To make the accepted regions easier to interpret, we intersect their boxes with segmentation masks. We use an off-the-shelf foreground segmentation model, BiRefNet~\citep{zheng2024birefnet}, in our ImageNet experiments. Other segmentation methods, such as SAM~\citep{sam}, could also be used when separate object masks are desired. The resulting intersections, shown as colored overlays, highlight the segmented content, while the boxes show the full regions tested by ablation. We do not perform further ablation tests on these intersections because relevant image cues may include object contours or surrounding context that the masks exclude. By inspecting these visualizations across images, we identify recurring patterns that help interpret concepts associated with each predicate.

We present the complete algorithm, ablation operator, and search settings in Appendix~\ref{sec:grounding_details}. We further validate our design choices by evaluating the attribution methods and comparing the guided search with a strong area-matched search baseline in Tables~\ref{tab:cam_grounding_comparison} and~\ref{tab:grounding-recovery}.

\section{Experiments}
\label{sec:exp}
We evaluate three complementary aspects of \textsc{VisionLogic}'s explanations. Section~4.1 analyzes the class-specific numerical predicates. Section~4.2 evaluates the usefulness of the concept explanations in human model-understanding tasks, and Section~4.3 examines representative visual concepts.

\subsection{Global Analysis of Feature Use}
\label{sec:global_predicate_structure}

\paragraph{Setup.}

We examine how features are activated, selected, and reused in ViT-B/16~\citep{dosovitskiy2021image}, ResNet-50~\citep{he2016deep}, ConvNeXt-Base~\citep{ConvNet}, and Swin-T~\citep{liu2021swin}. Our analysis distinguishes features whose predicates are active, represented by $A_c(x)$, from those selected to explain the prediction, represented by $Q_c(x)$ (Section~\ref{sec:predicates}). For each population $\mathcal D_c$ grouped by model-predicted class, we measure each predicate's activation and selection frequencies. We also examine feature reuse across classes and the number of active and selected predicates per image. Thresholds are fitted on source images and kept fixed for held-out evaluation.

\begin{figure*}[!b]
    \centering
    \includegraphics[width=\linewidth]{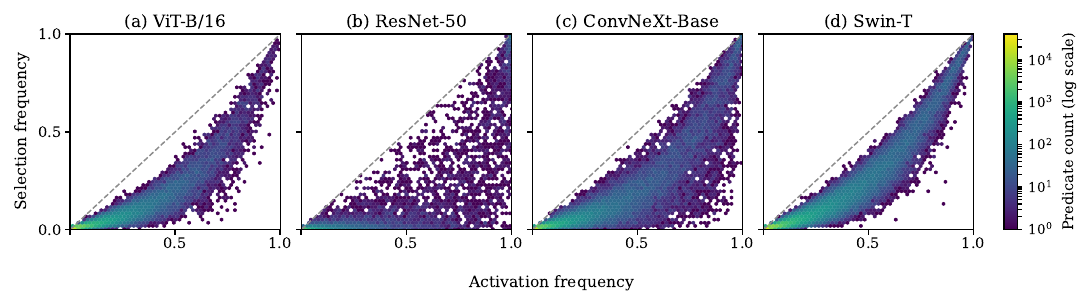}
    \vspace{-20pt}
    \caption{\textbf{Activation and selection frequencies on source images in four models.} Each hexbin counts class-specific predicates, with a shared logarithmic color scale. The dashed line indicates equal frequencies. Selection implies activation on source images by threshold construction. The large gaps below the line show that many frequently active features are selected much less often.}
    \label{fig:section42_activation_selection}
\end{figure*}

\begin{figure}[!t]
    \includegraphics[width=\linewidth]{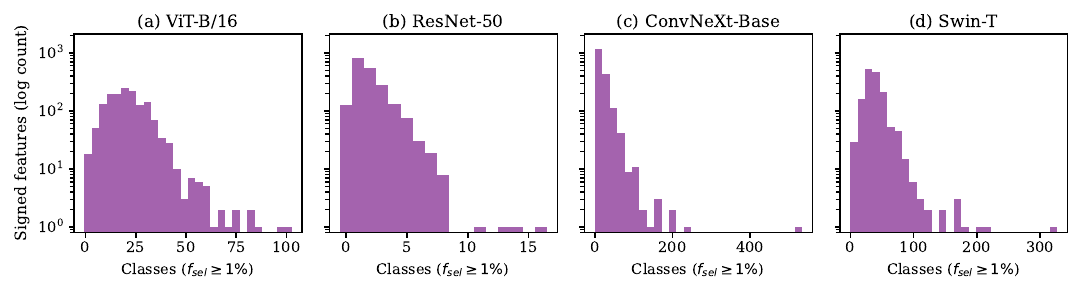}
    \vspace{-20pt}
    \caption{\textbf{Cross-class reuse of signed features in four models.} For each feature selected in the source images, we count the classes in which it is selected in at least 1\% of source examples. The vertical axis is shared and logarithmic; horizontal ranges differ across models. Predicates for shared features use class-specific thresholds, and feature indices are specific to each model.}
    \label{fig:section42_reuse}
    \par\medskip
    \includegraphics[width=\linewidth]{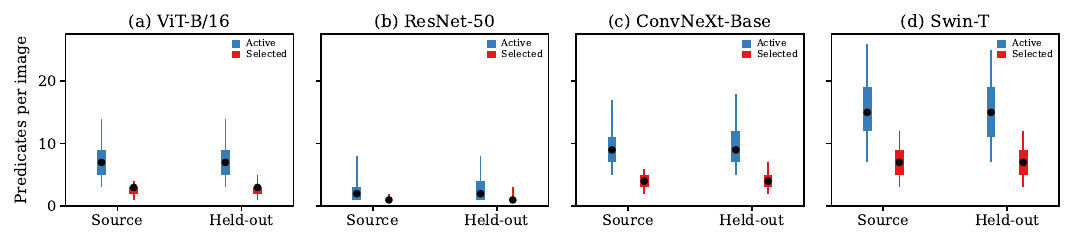}
    \vspace{-20pt}
    \caption{\textbf{Active and selected predicate counts per image.} For each model and data split, thin intervals span the 5th--95th percentiles, thick intervals show the interquartile range, and black points mark medians. These are distribution summaries, not confidence intervals.}
    \label{fig:section42_counts_appendix}
    \par\medskip
    \vspace{-10pt}
\end{figure}

\paragraph{Feature activation and selection.}
Figure~\ref{fig:section42_activation_selection} compares activation and selection frequencies within each predicted class. We observe that many predicates are active in a large proportion of source images, while their associated features are selected much less often. For example, in ViT class 326, predicate $p_{(257,+)}$ is active in 78.1\% of examples and selected in 10.0\%. In ConvNeXt class 713, the corresponding frequencies for predicate $p_{(90,-)}$ are 93.0\% and 9.3\%, respectively. On source images, selected features always satisfy their predicates because the thresholds are fitted from these activations. What stands out across all four models is how frequently a predicate remains active even when its feature is not selected. A feature can therefore be active in many images without often appearing in the compact sets that reproduce their predictions. This distinction helps separate common activation patterns from frequent use in our explanations.

\paragraph{Features shared across classes.}

Figure~\ref{fig:section42_reuse} shows how widely the same features are selected across different classes. Note that we count positive and negative activations of each feature separately within each model. For ConvNeXt, feature $(817,+)$ is selected in at least 1\% of source examples in 534 classes, and in at least 5\% of examples in 465 classes. For Swin, feature $(606,+)$ reaches the same cutoffs in 327 and 241 classes, respectively. Feature reuse varies considerably: some features are selected in only a few classes, while others recur across hundreds. These shared features connect explanations across classes, although their predicates use class-specific thresholds. Such reuse also raises the question of polysemanticity: does a feature represent one concept or different concepts across classes? Section~\ref{sec:eval_concept} explores this question through grounded visual examples.

\paragraph{Active and selected predicates per image.}

Figure~\ref{fig:section42_counts_appendix} summarizes how many predicates are active and selected for each image. The median active counts are 7, 2, 9, and 15 for ViT, ResNet, ConvNeXt, and Swin, respectively, while the corresponding median selected counts are 3, 1, 4, and 7. We apply the same feature selection procedure to source and held-out images. The median counts are unchanged between the two populations, and the interquartile ranges remain similar. Across all four models, the explanations typically involve only a few selected predicates, even when more are active. ResNet produces particularly compact explanations, with a median of one selected predicate per image. This means that one feature's contribution is often enough to reproduce the final head's prediction.

\paragraph{Predicates on held-out images.}
We examine whether predicates defined from the source images remain applicable to held-out images. Among all selected feature occurrences in held-out images, 98.6--99.5\% have a predicate defined for the same class and signed feature in the source data. Of these matched occurrences, 97.4--98.9\% satisfy the corresponding fixed threshold. Thus, most selected features on unseen images are covered by the source vocabulary and satisfy its thresholds without refitting. These strong generalization results support the simple feature-selection and predicate-construction procedure in Sections~\ref{sec:pathways}--\ref{sec:predicates}. Appendix~\ref{app:section42_details} reports the detailed results and the concentration of feature selections within each class.

\subsection{Human evaluation of ablation-grounded concepts}

\begin{table}[!t]
    \centering
    \caption{\textbf{Utility scores in three application scenarios.} The Utility benchmark measures how well explanations help users identify general rules that transfer to unseen instances. During training, participants are shown images along with the explanations and model predictions, and are asked to infer the underlying decision rules. At test time, the benchmark evaluates participants’ accuracy in predicting the model’s output on novel images. Higher Utility scores indicate that the explanations provide more useful information for understanding the model’s behavior on new samples. For each scenario, the first and second best results are in \textbf{bold} and \underline{underlined} respectively. \emph{\textsc{VisionLogic} achieves higher Utility scores than \textsc{ACE} and \textsc{CRAFT} across all three scenarios, with statistically significant improvements over both methods in the first two.}}
    \label{tab:human_eval}
    \resizebox{\textwidth}{!}{%
    \begin{tabular}{llrrrrrrrrrrrr}
        \toprule
        \multicolumn{2}{c}{} & \multicolumn{4}{c}{Husky vs. Wolf} & \multicolumn{4}{c}{Otter vs. Beaver} & \multicolumn{4}{c}{Kit Fox vs. Red Fox} \\
        \cmidrule(lr){3-6} \cmidrule(lr){7-10} \cmidrule(lr){11-14}
        & Session n$^{\circ}$ & 1 & 2 & 3 & Utility & 1 & 2 & 3 & Utility & 1 & 2 & 3 & Utility \\
        \midrule
        & Baseline & 65.7 & 68.6 & 70.3 & 1.00 & 84.4 & 90.3 & 92.2 & \underline{1.00} & 84.1 & 89.0 & 84.1 & \textbf{1.00} \\
        & Control & 55.3 & 63.6 & 70.0 & 0.92 & 85.1 & 88.3 & 92.9 & \underline{1.00} & 80.8 & 79.2 & 79.2 & 0.93 \\
        \midrule
        & \textsc{ACE} & 60.4 & 71.1 & 74.6 & \underline{1.01} & 80.4 & 85.7 & 90.5 & 0.96 & 80.6 & 83.2 & 76.2 & 0.93 \\
        & \textsc{CRAFT} & 55.5 & 60.8 & 65.3 & 0.89 & 86.3 & 90.9 & 90.9 & \underline{1.00} & 76.8 & 81.8 & 76.8 & 0.92 \\
        & \textsc{VisionLogic} & 74.8 & 90.0 & 91.0 & \textbf{1.25} & 96.8 & 98.4 & 99.2 & \textbf{1.10} & 84.1 & 84.5 & 82.9 & \underline{0.98} \\
        \bottomrule
    \end{tabular}%
    }
    \vspace{-7pt}
\end{table}

\begin{figure*}[!b]
    \centering
    \includegraphics[width=\linewidth]{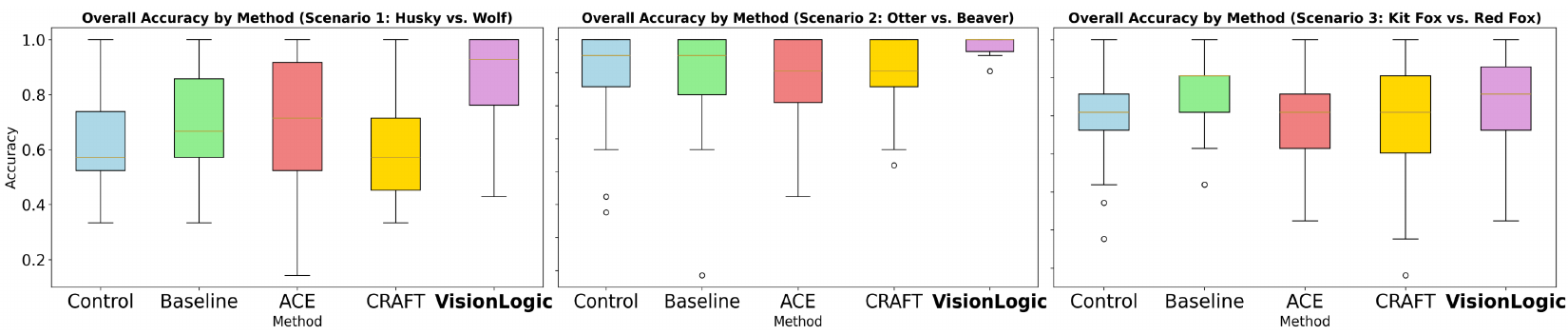}
    \vspace{-10pt}
    \caption{\textbf{Participant accuracy distributions for each explanation method.} \textsc{VisionLogic} achieves higher mean accuracy than the \textsc{ACE} and \textsc{CRAFT} baselines across all three scenarios.}
    \vspace{-10pt}
    \label{fig:boxplots}
\end{figure*}



\paragraph{Setup.}
We evaluate the practical utility of our proposed \textsc{VisionLogic}, alongside prior concept-based methods \textsc{ACE} and \textsc{CRAFT}, in terms of \emph{concept explanations} using the human-in-the-loop framework of \citet{DBLP:conf/nips/ColinFCS22}, which assesses how well explanations help participants understand a model's behavior across three real-world scenarios: (1) detecting bias in AI decisions (using Husky vs. Wolf classification), (2) identifying novel model strategies that are non-obvious to untrained observers (using Otter vs. Beaver classification), and (3) understanding failure cases (using Kit Fox vs. Red Fox classification). Each scenario adopts the meta-predictor paradigm: during the training phase, participants study example images paired with explanation images and model outputs, then predict the model’s output on unseen images without access to the corresponding explanations.

In total, 531 participants were recruited from Prolific \citep{prolific}, of whom 465 passed screening criteria aligned with \citet{DBLP:conf/nips/ColinFCS22}. We designed questionnaires for five conditions: (1) baseline (no explanation), (2) control (bottom-up saliency maps~\citep{SimonyanVZ13}), (3) \textsc{ACE}~\citep{ace}, (4) \textsc{CRAFT}~\citep{craft}, and (5) \textsc{VisionLogic} (ours), where \textsc{ACE} and \textsc{CRAFT} are considered state-of-the-art. To ensure fair comparison on the adapted datasets, all methods were re-run under identical experimental settings. Following prior work \citep{DBLP:conf/nips/ColinFCS22,craft}, we report the \emph{utility score}, defined as participants’ average test accuracy across the three sessions, normalized by the baseline; higher values indicate more effective human understanding of the model. 
Additional experimental details, including the participant recruitment process and the number of images used in each session and phase, are provided in Appendix~\ref{appdx: human_eval}.

\begin{figure*}[!b]
    \centering
    \includegraphics[width=0.94\linewidth]{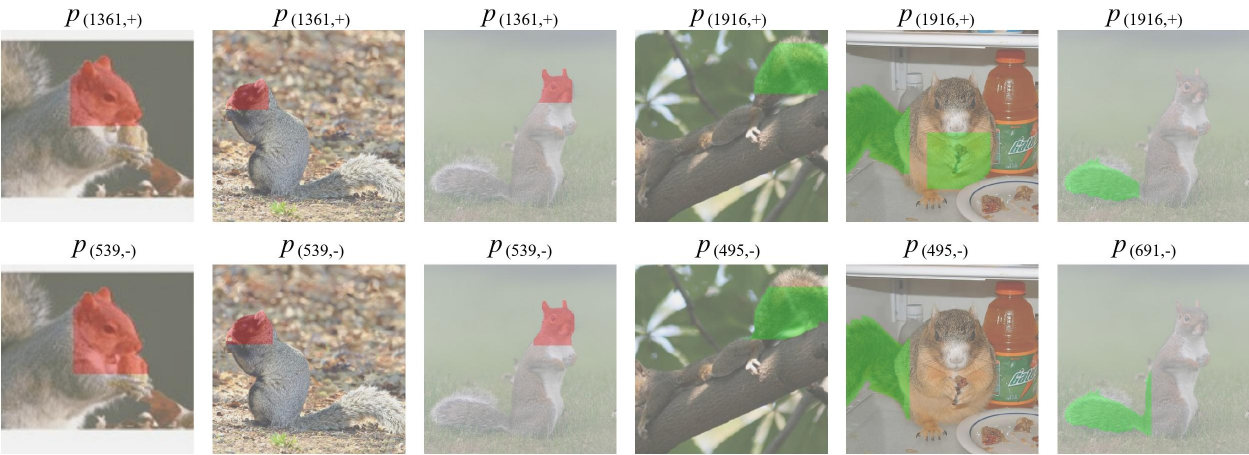}
    \includegraphics[width=0.94\linewidth]{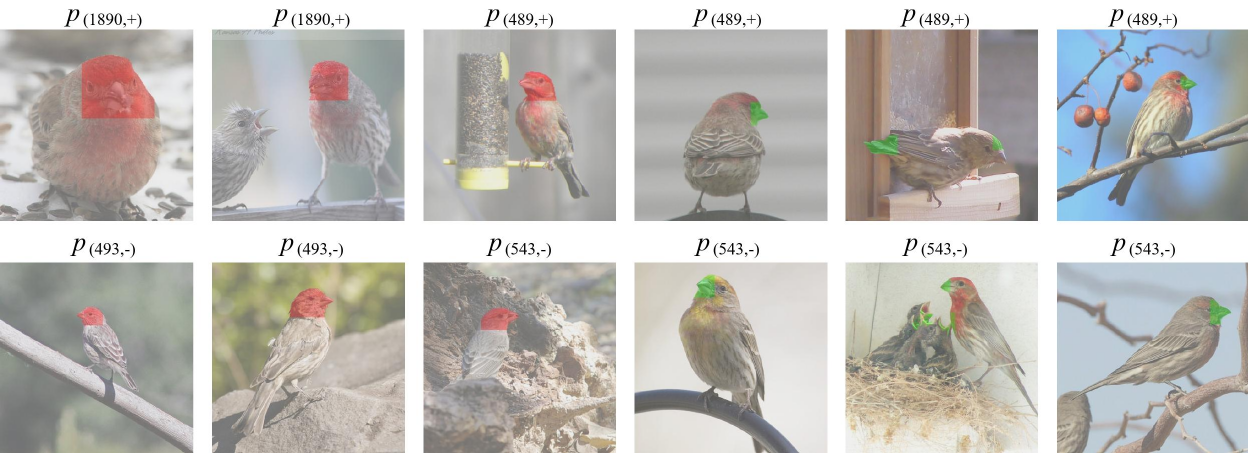}
    \includegraphics[width=0.94\linewidth]{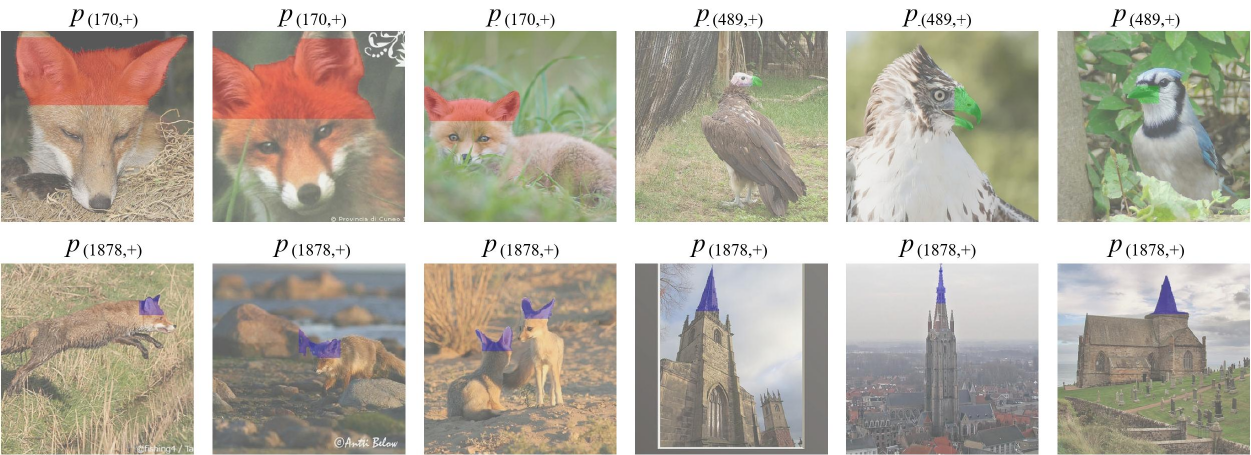}

    \caption{\textbf{Human-recognizable visual concepts grounded through predicates.} In each image, the predicate appears above the frame, and the colored overlay highlights the associated visual content.}
    \label{fig:masked_images}
\end{figure*}

    
    
    
    













\paragraph{Results.}
Table~\ref{tab:human_eval} shows that \textsc{VisionLogic} achieves higher utility than \textsc{ACE} and \textsc{CRAFT} across three scenarios, with statistically significant gains over both in the first two. Scores of 1.25 and 1.10 represent improvements of 25\% and 10\% over the no-explanation baseline, respectively. In the third scenario, we observe the same trend as prior work \citep{craft}, where no existing method provides more effective information than Baseline. Nevertheless, \textsc{VisionLogic} achieves the highest utility among the evaluated explanation methods, approaching the no-explanation baseline at 0.98.

To rigorously examine the performance gain of our method over Baseline, Control, \textsc{ACE}, and \textsc{CRAFT} in the evaluation shown in Table~\ref{tab:human_eval}, we perform statistical tests on the collected data. We initially follow the procedure of \citet{DBLP:conf/nips/ColinFCS22}: an analysis of variance (ANOVA; \citealp{scheffe1999analysis}) followed by Tukey’s honestly significant difference (HSD) test for pairwise comparisons \citep{tukey1949comparing}. However, we noticed that the assumption checks for normality and homogeneity, which are crucial to the validity of parametric test results \citep{oztuna2006investigation}, are absent from prior works \citep{craft}. To this end, we performed a complete statistical testing procedure with assumption checks. Figure~\ref{fig:boxplots} shows that the data is highly skewed as test accuracy is upper bounded by 1.0, hence violating the normality assumption. Therefore, non-parametric statistics are more appropriate for testing whether there is any statistically significant difference between the five conditions listed.

A Kruskal-Wallis test \citep{mckight2010kruskal} with a null hypothesis of \textit{``All condition distributions are identical''} and an alternative hypothesis of \textit{``At least one condition distribution differs''} at a significance level of 0.05 rejects the null hypothesis with $p=3.4\times10^{-5},\ 8.83\times10^{-3}$ for the first two scenarios. We then utilize Dunn's test \citep{dinno2015nonparametric} to analyze pairwise differences with Bonferroni correction \citep{weisstein2004bonferroni} applied. In the first scenario, \textsc{VisionLogic} is shown to be \emph{significantly} better than all other conditions with $p = 3.03\times10^{-2},\ 4.00\times10^{-4},\ 2.41\times10^{-2},\ p<0.001$, respectively, suggesting that our method is effective in helping participants detect biases in the model. In the second scenario, \textsc{VisionLogic} is shown to be \emph{significantly} more effective than \textsc{ACE} and \textsc{CRAFT} with $p=4.5\times10^{-3} \text{ and } 3.09\times10^{-2}$ respectively, supporting its improvement over prior methods on the task of identifying non-obvious visual cues. Full test details, including the statistical tests involved, assumption checks, corrections, and test statistics are reported in Appendix~\ref{appdx:human_eval_stats}.

\subsection{Qualitative analysis of visual concepts}
\label{sec:eval_concept}

We present human-interpretable visual concepts encoded by predicates discovered in both ResNet and ViT. Each concept judgment is based on consistent visual inspection across many instances; Figure~\ref{fig:masked_images} shows representative cases. The 2\textsuperscript{nd} and 4\textsuperscript{th} rows are from ViT, the others from ResNet. Sampled concepts include \textit{squirrel head}, \textit{squirrel tail/paws}, \textit{bird head}, \textit{bird beak}, \textit{fox ears}, and \textit{church tops}.

\paragraph{Polysemanticity.}
We observe a many-to-many relationship between predicates and concepts (polysemanticity). Within a class, regions associated with one predicate can contain different visual concepts; across classes, identical abbreviated labels indicate reuse of the same (signed) feature. Conversely, regions associated with different predicates can depict the same concept. We interpret these recurring visual associations as evidence of polysemanticity, with ablation validating the combined region. For example, in the 5\textsuperscript{th} image of row~1, both the tail and paws of the squirrel are captured by $p_{(1916,+)}$ (ResNet). Predicate $p_{(1878,+)}$ simultaneously encodes \textit{fox ears} and \textit{church tops}; while semantically different, both share a triangular geometry, which may explain the reuse. Predicate $p_{(489,+)}$ (rows~3 and~5) frequently encodes \textit{bird beak} across species (\textit{jay}, \textit{vulture}, \textit{kite}, etc.), and recurs in local explanations for bird images, indicating an influential role in inference.

We also find cases where a single concept is captured by multiple predicates. For the \textit{fox} examples (rows~5--6), both $p_{(1878,+)}$ and $p_{(170,+)}$ encode \textit{fox ears}. A predicate may also attend to multiple instances of the same concept within an image: in row~6, image~3 (two foxes), $p_{(1878,+)}$ captures both pairs of ears. Such behaviors highlight the role of predicates as global concept detectors rather than local ones.

\paragraph{CNNs vs.\ Transformers.}

Similar concepts are found in both model families (see rows 1--2, where both capture the same concept on the same image). In these examples, ViT tends to involve \emph{more} predicates per visual concept, whereas ResNet yields sparser, more distinct encodings. These observations complement the quantitative analysis of feature use in Section~\ref{sec:global_predicate_structure}. We hypothesize two factors behind this difference: (1) activation functions (ReLU vs.\ GELU) induce different predicate sparsity through positive and negative activation branches; and (2) convolution versus attention imposes different inductive biases on concept formation. Validating these hypotheses and examining how architectural differences shape the discovered concepts remain directions for future work.

\section{Conclusion}
\label{sec:conclusion}

Existing concept-based explanation methods for vision models often rely on \emph{correlational evidence}, leaving the connection between recognizable concepts and prediction-relevant internal features insufficiently grounded. To address this gap, we propose \textsc{VisionLogic}, a post-hoc framework that grounds these features in visual concepts through intervention-based validation. Beyond explaining individual predictions, the same predicates allow us to examine how features are activated, selected, and reused across images and classes. Experiments on CNNs and ViTs show that compact sets of features can reproduce a prediction even when many more features are active. They also reveal extensive feature reuse across classes. Our large-scale human evaluation with 465 participants further demonstrates the practical value of the discovered concepts: \textsc{VisionLogic}'s explanations achieve higher Utility scores than \textsc{ACE} and \textsc{CRAFT} across three model-understanding tasks. We envision this work as a step toward bridging the gap between recognizable visual concepts and an understanding of how vision models make decisions.

\paragraph{Limitations.}
Our grounding procedure uses localized blur interventions that preserve image content outside the candidate regions. However, these interventions can still introduce distribution shift, so changes in feature activation may reflect intervention artifacts as well as the removal of relevant visual evidence. Distinguishing these effects remains an open challenge in explainable AI.

\label{maintext:end}
\clearpage

\bibliography{iclr2027_conference}
\bibliographystyle{iclr2027_conference}

\appendix
\clearpage
\newpage

\section{Experimental setup}
\label{sec:setup}

\paragraph{Models and data.}
We study feature activation, selection, and reuse in four pretrained ImageNet-1k models: ResNet-50~\citep{he2016deep}, ConvNeXt-Base~\citep{ConvNet}, ViT-B/16~\citep{dosovitskiy2021image}, and Swin-T~\citep{liu2021swin}. For each model, we select compact sets of features whose contributions through the final classification head reproduce its predictions. We then use the selected source activations to fit class-specific thresholds and apply these thresholds to held-out images without refitting. Section~\ref{sec:global_predicate_structure} presents the main findings, and Appendix~\ref{app:section42_details} gives the image populations and evaluation metrics. The following sections explain how we select features, construct and ground predicates, analyze feature use, and evaluate the resulting explanations with human participants.

\paragraph{Hardware.}
We ran the experiments on Ubuntu~22.04 LTS with an AMD EPYC\textsuperscript{TM} 7532 processor (32 cores), 128~GB RAM, and a single NVIDIA A100 GPU with 40~GB memory. The pretrained vision models are kept fixed throughout feature selection, predicate construction, and visual grounding.

\section{Feature selection and predicate implementation}
\label{app:method_implementation}


\paragraph{Source image selection.}
We use the ImageNet-1k training split as the source population and the ImageNet
validation split as held-out data. For each class, we use its correctly
classified training images as the source population for computing the numerical
contributions and predicates described in
Sections~\ref{sec:pathways}--\ref{sec:predicates}.
Appendix~\ref{app:section42_details} gives the resulting source and held-out
populations.

\begin{algorithm2e}[!ht]
\small
\DontPrintSemicolon
\SetKw{Return}{return}
\SetKw{Break}{break}
\SetKwProg{Fn}{Procedure}{:}{}
\caption{Feature selection, predicate construction, and held-out evaluation}
\label{alg:predicate-extraction}
\KwIn{Frozen representation and classification head $(z,W,b)$; retained source images; shared stable sorting convention.}
\KwOut{Ordered selected features $S$; class-specific predicate vocabularies and thresholds $\{\mathcal V_c,T_c\}$; active and selected sets for held-out images.}
\Fn{\textsc{SelectFeatures}$(x)$}{
 $c\leftarrow\arg\max_k (Wz(x)+b)_k$\;
 $\pi\leftarrow$ decreasing contribution order of $W_{cj}z_j(x)$\;
 $\ell\leftarrow b$; $S\leftarrow()$\;
 \For{$r=1,\ldots,d$}{
  $j\leftarrow\pi_r$; $s\leftarrow+$ if $z_j(x)\geq0$, otherwise $-$\;
  $\ell\leftarrow\ell+W_{:j}z_j(x)$; append $(j,s)$ to $S$\;
  \If{$\arg\max_k\ell_k=c$}{\Break}
 }
 \Return $(c,S)$\;
}
\ForEach{retained source image $x$}{
 $(c,S(x))\leftarrow\textsc{SelectFeatures}(x)$; store the predicted class $c$\;
}
Initialize $\mathcal V_c\leftarrow\varnothing$ for each source class\;
\ForEach{retained source tuple $(x,c,S(x))$}{
 \ForEach{$(j,s)\in S(x)$}{
  \If{$(j,s)\notin\mathcal V_c$}{
   $T_{c,(j,s)}\leftarrow+\infty$ for $s=+$, otherwise $-\infty$\;
   $\mathcal V_c\leftarrow\mathcal V_c\cup\{(j,s)\}$\;
  }
  $T_{c,(j,s)}\leftarrow\min(T_{c,(j,s)},z_j(x))$ for $s=+$; otherwise use $\max$\;
 }
}
Freeze all $\mathcal V_c$ and $T_{c,(j,s)}$\;
\ForEach{held-out image $x$}{
 $(c,S(x))\leftarrow\textsc{SelectFeatures}(x)$\;
 \If{$\mathcal V_c$ is unavailable}{continue\;}
 $A_c(x)\leftarrow\{(j,s)\in\mathcal V_c:p_{c,(j,s)}(x)=1\}$\;
 $Q_c(x)\leftarrow\operatorname{set}(S(x))\cap\mathcal V_c$\;
 $J_c(x)\leftarrow A_c(x)\cap Q_c(x)$\;
 Record $(c,S(x),A_c(x),Q_c(x),J_c(x))$ and selected features outside $\mathcal V_c$\;
}
\end{algorithm2e}


\paragraph{Feature ordering and numerical implementation.}
We order features by decreasing contribution to the predicted class using \texttt{np.argsort(contributions, kind='stable')[::-1]}. The same convention is used for all four models and for both source and held-out images, giving a consistent ordering when contributions are equal. Predictions are obtained with \texttt{np.argmax}, which chooses the first class index in a tie. We apply the stopping condition directly, without adding a tolerance or margin. The positive activation branch includes zero; negative features $(j,-)$ are stored as $j+d$.

The full prediction is computed from \texttt{matrix @ weights.T + bias}. To find a compact set of features that reproduces it, we begin with \texttt{bias.copy()} and add each feature's contribution to all class logits in order (Section~\ref{sec:pathways}). Matrix multiplication and sequential addition can produce small floating-point differences. We access the final classification heads through \texttt{heads.head} in ViT, \texttt{fc} in ResNet, \texttt{classifier.2} in ConvNeXt, and \texttt{head} in Swin.

\paragraph{Feature selection and predicate construction.}
For each source image, we identify a compact set of features whose contributions through the frozen classification head reproduce its prediction. We record these features in contribution order as $S(x)$ and use their activations to fit the class-specific thresholds. Threshold fitting uses exactly this selection, ending at the first prefix that recovers the prediction. Every selected source sequence reproduces the original head prediction. Features shared across classes contribute to separate class-specific predicates, allowing their activation thresholds to differ even when the selected sequences are identical.

\paragraph{Held-out evaluation.}
For a held-out image, we select features using the same procedure and evaluate the predicates fitted for its predicted class (Algorithm~\ref{alg:predicate-extraction}). Their thresholds remain fixed, allowing us to examine how well the source predicates apply to unseen images. We obtain the active set $A_c(x)$ and selected set $Q_c(x)$, then ground the predicates associated with their intersection $J_c(x)$ in visual regions.

\section{Grounding predicates in visual concepts}
\label{sec:grounding_details}

\paragraph{Input and predicate selection.}
We resize the shorter side of each image to 256 pixels using bilinear interpolation and take a $224\times224$ center crop. Standard ImageNet normalization is applied before classifier evaluation. Given an image $x$, we fix its original prediction $c=\hat y(x)$ and consider the predicates associated with $J_c(x)=A_c(x)\cap Q_c(x)$ (Section~\ref{sec:predicates}). These predicates are active on the image, and their features are selected to explain its prediction. Their class-specific thresholds are fitted on the source images and remain fixed during grounding. We examine each predicate independently to identify the visual regions supporting its activation. Selected features without an active source predicate remain in the numerical explanation but are not grounded visually.

\paragraph{Neuron-specific attribution maps.}
To locate candidate regions, we use attribution maps that target the signed activation $s z_j$ supplied to the final classification head, with $s=1$ for the positive branch and $s=-1$ for the negative branch. The maps therefore highlight regions associated with the feature we aim to ground. For Grad-CAM~\citep{selvaraju2017gradcam}, let $A_k$ be channel $k$ at the chosen spatial layer and $\Omega$ its spatial index set. We compute
\begin{equation}
\gamma_k=\frac{1}{|\Omega|}\sum_{u\in\Omega}\frac{\partial(s z_j)}{\partial A_k(u)},
\qquad
H_{\mathrm{grad}}=\mathcal N\!\left(\operatorname{ReLU}\!\left(\sum_k\gamma_k A_k\right)\right).
\label{eq:neuron-gradcam}
\end{equation}
If the Grad-CAM search fails to find an accepted region, we repeat the search with Score-CAM~\citep{wang2020scorecam}. Let $M_k$ be the bilinearly upsampled and spatially min--max-normalized map of $A_k$, and $\bar x$ the normalized classifier input. We compute
\begin{equation}
a_k=s z_j(\bar x\odot M_k),\qquad
\omega_k=\frac{\exp(a_k)}{\sum_\ell\exp(a_\ell)},\qquad
H_{\mathrm{score}}=\mathcal N\!\left(\operatorname{ReLU}\!\left(\sum_k\omega_k A_k\right)\right),
\label{eq:neuron-scorecam}
\end{equation}
where $z_j$ in Equation~\ref{eq:neuron-scorecam} is evaluated on the masked normalized input, and $\mathcal N$ denotes spatial normalization and resizing to the input grid. For negative predicates, both methods target decreasing $z_j$.

We follow the \texttt{pytorch-grad-cam} implementation conventions. Score-CAM evaluates masks from all spatial channels in batches of eight, sharing these forward evaluations across neuron targets. A softmax over the signed neuron scores gives the channel weights. The weighted spatial sum is then clipped at zero, normalized, resized, and normalized again to produce the heatmap. Input masks use bilinear interpolation with \texttt{align\_corners=True} and a $10^{-8}$ denominator stabilizer. Both attribution methods provide candidate regions for the subsequent ablation tests.

We obtain spatial activations from \texttt{layer4.2} in ResNet-50, \texttt{features.7} in ConvNeXt-Base, \texttt{encoder.layers.encoder\_layer\_11.ln\_1} in ViT-B/16, and \texttt{features.7.1.norm1} in Swin-T. For ViT, we remove the class token and reshape the remaining tokens into a $14\times14$ grid. For Swin, we permute channels-last features to channels-first. These spatial layers provide the maps used for localization, while the attribution target is the signed activation of the selected feature at the final classification head. The chosen layer determines the heatmap's spatial resolution.

\paragraph{Iterative region refinement and ablation.}
Algorithm~\ref{alg:pred_region} gives the grounding procedure described in Section~\ref{sec:grounding}. We first propose boxes around areas with strong Grad-CAM responses. If ablating these regions leaves the predicate active, we progressively include weaker responses and test the resulting boxes. When the complete Grad-CAM search fails, we repeat the procedure with Score-CAM. We accept the first union of boxes whose ablation deactivates the predicate. If neither search succeeds, the predicate remains unlocalized for that image.

Let $x$ be the original $224\times224$ RGB crop, $B$ a candidate union of boxes, and $G_{20}(x)$ the image blurred with a Gaussian blur radius of 20 pixels. We ablate $B$ by replacing its content with the corresponding blurred pixels:
\begin{equation}
x^{\setminus B}=(1-\mathbf{1}_B)\odot x+\mathbf{1}_B\odot G_{20}(x).
\label{eq:grounding-blur}
\end{equation}
Throughout the search, we keep the original class, activation sign, and threshold fixed. A nonempty candidate is accepted when
\begin{equation}
p_{c,(j,s)}(x)=1
\quad\text{and}\quad
p_{c,(j,s)}(x^{\setminus B})=0.
\label{eq:grounding-acceptance}
\end{equation}

\begin{algorithm2e}[!hb]
    \caption{Iterative visual grounding with neuron-specific attribution and ablation tests}
    \label{alg:pred_region}
    \small
    \DontPrintSemicolon
    \KwIn{RGB image $x$; predicate $p=p_{c,(j,s)}$ with $(j,s)\in J_c(x)$; fixed class, sign, and threshold}
    \KwOut{Accepted box union $B$ and foreground overlay mask $D$, or \textsc{NoRegion}}
    $x_{\mathrm{blur}}\leftarrow\textsc{GaussianBlur}(x,\mathrm{radius}=20)$\;
    \ForEach{$m\in(\textsc{GradCAM},\textsc{ScoreCAM})$}{
        $H\leftarrow\textsc{SignedNeuronAttribution}(m,x,j,s)$\;
        \ForEach{$\alpha\in(0.60,0.55,\ldots,0.10)$}{
            $U_\alpha\leftarrow\{u:H(u)\geq\alpha\max(H)\}$ if $\max(H)>0$, otherwise $\varnothing$\;
            $\mathcal C_\alpha\leftarrow\textsc{ConnectedComponents}(U_\alpha,\mathrm{connectivity}=8)$\;
            $B\leftarrow\bigcup_{C\in\mathcal C_\alpha}\textsc{TightBox}(C)$\;
            \If{$B\neq\varnothing$}{
                $x'\leftarrow(1-\mathbf{1}_B)\odot x+\mathbf{1}_B\odot x_{\mathrm{blur}}$\;
                \If{$p(x')=0$}{
                    $F\leftarrow\{u:\textsc{BiRefNet}(x)(u)\geq0.5\}$\;
                    $D\leftarrow B\cap F$ \tcp*{Foreground visualization}
                    \Return $(B,D)$\;
                }
            }
        }
    }
    \Return \textsc{NoRegion}\;
\end{algorithm2e}

For each attribution map, we use eleven cutoffs $\alpha=0.60,0.55,\ldots,0.10$, relative to the maximum heatmap response. At each cutoff, we find all nonempty 8-connected components and draw tight axis-aligned boxes around them. We retain small components as well as large ones and combine their boxes into a single ablation mask. The boxes have no added margins or foreground restriction. A zero heatmap provides no candidate regions for that attribution method.

As the cutoff decreases, weaker responses enter the candidate region, potentially enlarging or merging its boxes. A larger region need not produce a larger change in the target activation, so we recompute the predicate after each ablation. All boxes for one predicate are tested together, and each predicate is examined separately using the original image. Once a region passes the test, we retain it without further area minimization or random restarts.

We compute \texttt{PIL.ImageFilter.GaussianBlur(radius=20)} once on the original RGB crop and use the corresponding blurred pixels for every candidate region. The radius is measured at the $224\times224$ input resolution; it is not a 20-pixel kernel width. Each ablation starts from the original image and is applied before model normalization, leaving pixels outside the boxes unchanged. A positive predicate becomes inactive when $z_j(x')<T_{c,(j,+)}$, while a negative predicate becomes inactive when $z_j(x')>T_{c,(j,-)}$. Equality remains active. This threshold crossing determines acceptance, without requiring a change in the predicted class or a separate amount of activation suppression.

Blur removes fine detail while preserving some low-frequency structure. It can therefore leave relevant cues intact or introduce artifacts at region boundaries. When a search fails, we leave the predicate unlocalized because none of the tested regions deactivated it under the chosen blur operator and cutoff schedule.

\paragraph{Foreground visualization.}
We use the pretrained \texttt{ZhengPeng7/BiRefNet} model~\citep{zheng2024birefnet} to highlight the foreground content within accepted regions. We resize the original crop to $1024\times1024$ and apply ImageNet normalization before segmentation. A sigmoid converts the final output to foreground probabilities, which are resized bilinearly to $224\times224$ with \texttt{align\_corners=False} and thresholded at $0.5$. We use this mask directly, without output min--max normalization or dilation. Its intersection with the accepted boxes gives the foreground overlay, without further restriction to the heatmap. An empty intersection produces no overlay but leaves the accepted boxes available for inspection.

We provide two complementary views of each explanation. Solid colored boundaries show the full boxes tested by ablation, while colored overlays on a lightened image highlight their foreground content. Regions associated with the same predicate share a color, including disconnected boxes, and different predicates receive different colors. Where overlays overlap, their colors are averaged. Since segmentation depends only on the image, its mask can be computed once and reused for all predicates.

Foreground segmentation makes the accepted regions easier to interpret, as described in Section~\ref{sec:grounding}. We do not apply further ablation tests to the intersections because relevant cues may include object contours or surrounding context excluded by the masks. The box view preserves this context, while the overlays help readers identify recurring visual patterns and interpret the concepts associated with each predicate.

\paragraph{Grounding evaluation setup.}
We evaluate the attribution methods and the combined guided search on the same seed-42 sample of 100 ImageNet validation images for ViT-B/16 and ResNet-50. The evaluation includes all predicates associated with $J_c(x)$: 276 predicate occurrences across 99 eligible ViT images and 129 across 100 ResNet images. All methods use the same heatmap cutoffs, connected-region box construction, and predicate-deactivation test described above.

We report predicate success, the fraction of tested predicate occurrences for which an accepted region is found, and image coverage, the fraction of eligible images with at least one accepted region. To assess the size of the resulting explanations, we also report occupancy, the fraction of image area covered by an accepted box union. Occupancy is summarized over successful predicate tests.

\paragraph{Attribution methods and search order.}
Table~\ref{tab:cam_grounding_comparison} compares Grad-CAM, Score-CAM, and their sequential combination. Grad-CAM produces more compact accepted regions, with median occupancy of 23.0\% for ViT and 41.2\% for ResNet, compared with 33.1\% and 57.9\% for Score-CAM. Score-CAM, however, succeeds on more predicates. Their combination achieves the highest predicate success and image coverage among the evaluated methods, while keeping median occupancy closer to Grad-CAM.

\begin{table}[!t]
    \centering
    \caption{\textbf{Comparison of attribution methods for visual grounding.} All methods use the same 100-image sample per model and predicate-deactivation test. Image coverage is measured over images containing at least one eligible predicate. Median occupancy is computed over successful predicate tests. The combined search applies Grad-CAM first and Score-CAM only when the Grad-CAM search fails.}
    \label{tab:cam_grounding_comparison}
    \resizebox{\linewidth}{!}{%
    \begin{tabular}{llccc}
        \toprule
        Model & Attribution method & Predicate success & Image coverage & Median occupancy \\
        \midrule
        ViT-B/16
        & Grad-CAM
        & $242/276$ ($87.7\%$)
        & $98/99$ ($99.0\%$)
        & $23.0\%$ \\
        & Score-CAM
        & $254/276$ ($92.0\%$)
        & $97/99$ ($98.0\%$)
        & $33.1\%$ \\
        & Grad-CAM $\rightarrow$ Score-CAM
        & $\mathbf{274/276}$ ($\mathbf{99.3\%}$)
        & $\mathbf{99/99}$ ($\mathbf{100.0\%}$)
        & $24.0\%$ \\
        \midrule
        ResNet-50
        & Grad-CAM
        & $95/129$ ($73.6\%$)
        & $79/100$ ($79.0\%$)
        & $41.2\%$ \\
        & Score-CAM
        & $109/129$ ($84.5\%$)
        & $86/100$ ($86.0\%$)
        & $57.9\%$ \\
        & Grad-CAM $\rightarrow$ Score-CAM
        & $\mathbf{116/129}$ ($\mathbf{89.9\%}$)
        & $\mathbf{90/100}$ ($\mathbf{90.0\%}$)
        & $46.3\%$ \\
        \bottomrule
    \end{tabular}%
    }
\end{table}

Score-CAM recovers 32 of 34 failed Grad-CAM predicate tests for ViT and 21 of 34 for ResNet, showing that the two methods provide complementary proposals. These results support our choice to begin with Grad-CAM, which requires one backward pass and supplies most accepted regions, and invoke Score-CAM when needed. This ordering improves the availability of grounded explanations while avoiding the additional Score-CAM computation whenever Grad-CAM succeeds.

\paragraph{Comparison with area-matched search.}
To examine the benefit of attribution-guided proposals, we compare the combined search with a strong area-matched search baseline. At each cutoff, the baseline randomly places a rectangle with approximately the same area and enclosing aspect ratio as the guided box union. It mirrors both stages of the guided search and succeeds if either corresponding search finds a region that deactivates the predicate. Matching proposal area, search opportunities, and the acceptance test makes this a controlled comparison of the benefit of attribution-based spatial guidance.

\begin{table}[!t]
    \centering
    \caption{\textbf{Guided grounding versus area-matched search on 100 images per model.} The baseline mirrors both stages of the guided search with approximately area-matched proposals placed at random locations. Image coverage refers to the guided search. Occupancy is summarized over successful predicate tests as the median percentage of image area covered; brackets give the interquartile range.}
    \label{tab:grounding-recovery}
    \resizebox{\linewidth}{!}{%
    \begin{tabular}{lccccc}
        \toprule
        Model & Guided success & Baseline success & Image coverage & Guided occupancy & Baseline occupancy \\
        \midrule
        ViT-B/16
        & $274/276$ (99.3\%)
        & $256/276$ (92.8\%)
        & $99/99$ (100\%)
        & 24.0\% [11.8, 50.0]
        & 27.9\% [16.8, 56.4] \\
        ResNet-50
        & $116/129$ (89.9\%)
        & $105/129$ (81.4\%)
        & $90/100$ (90.0\%)
        & 46.3\% [24.8, 70.8]
        & 60.7\% [37.0, 78.7] \\
        \bottomrule
    \end{tabular}%
    }
\end{table}

For both architectures, the guided search deactivates more predicates than the area-matched baseline (Table~\ref{tab:grounding-recovery}). Its accepted regions also have lower median occupancy: 24.0\% versus 27.9\% for ViT and 46.3\% versus 60.7\% for ResNet. These results support our use of neuron-specific attribution to guide iterative refinement toward visual regions that pass the ablation test.

\section{Additional analysis of feature use}
\label{app:section42_details}

This appendix provides the definitions, image populations, and additional results for the analysis in Section~\ref{sec:global_predicate_structure}. For each model, $p_{c,(j,s)}$ represents the activation state of signed feature $(j,s)$ using a threshold $T_{c,(j,s)}$ fitted for predicted class $c$. Positive predicates are active at or above the minimum selected source activation, while negative predicates are active at or below the corresponding maximum. We apply the same feature selection procedure to source and held-out images, selecting compact sets whose contributions reproduce the original prediction. Thresholds are fitted on the selected source activations and remain fixed for held-out evaluation. Each image is analyzed using the predicates defined for its predicted class. The same features may be selected across different classes, with separate thresholds describing their activation states in each class.

We select source images using the correct-classification procedure described in
Appendix~\ref{app:method_implementation}. Table~\ref{tab:section42_populations} summarizes the resulting populations for each model. Held-out evaluation uses all 50,000 ImageNet validation images, grouped by model prediction. ConvNeXt has no source predicates for class 744, so the 34 validation images predicted as this class are excluded from analyses requiring a source vocabulary. These analyses therefore use 49,966 ConvNeXt images. We compare distributions across models, with each model evaluated on its own predicted-class groups.
\FloatBarrier

\begin{table}[!t]
    \caption{\textbf{Image populations and predicate counts.} Dim.\ gives the number of features supplied to the classification head, and Vocab.\ gives the median number of source predicates per predicted class. Source and Held-out report median active/selected predicate counts per image. Models follow the order used in the main figures.}
    \label{tab:section42_populations}
    \centering
    \small
    \setlength{\tabcolsep}{3.2pt}
    \begin{tabular}{lrrrrr}
        \toprule
        Model & Dim. & Source images & Vocab. & Source & Held-out \\
        \midrule
        ViT-B/16      & 768   & 1,206,951 & 95  & 7/3  & 7/3 \\
        ResNet-50     & 2,048 & 1,169,769 & 7   & 2/1  & 2/1 \\
        ConvNeXt-Base & 1,024 & 1,216,462 & 81  & 9/4  & 9/4 \\
        Swin-T        & 768   & 1,153,631 & 133 & 15/7 & 15/7 \\
        \bottomrule
    \end{tabular}
\end{table}
\FloatBarrier

\paragraph{Activation and selection frequencies.}
Let $\mathcal D_c$ contain the source or held-out images assigned to predicted class $c$, with $|\mathcal D_c|>0$. For each signed feature $(j,s)\in\mathcal V_c$, we measure its predicate's activation frequency and its feature's selection frequency:
\begin{equation}
 f_{\mathrm{act}}(c,j,s)=\frac{\sum_{x\in\mathcal D_c}p_{c,(j,s)}(x)}{|\mathcal D_c|},
 \qquad
 f_{\mathrm{sel}}(c,j,s)=\frac{\sum_{x\in\mathcal D_c}\mathbf{1}\{(j,s)\in S(x)\}}{|\mathcal D_c|}.
 \label{eq:predicate-frequency}
\end{equation}
These frequencies describe the proportion of all images in the class for which the predicate is active or its feature is selected. We obtain $S(x)$ using the same feature selection procedure for both populations and evaluate predicates with their fixed source thresholds. For cross-class reuse, we count how many classes meet a given selection-frequency cutoff for the same feature $(j,s)$. Positive and negative activations are counted separately, and the predicates associated with a shared feature retain their class-specific thresholds.

\begin{figure}[!hb]
    \centering
    \includegraphics[width=\linewidth]{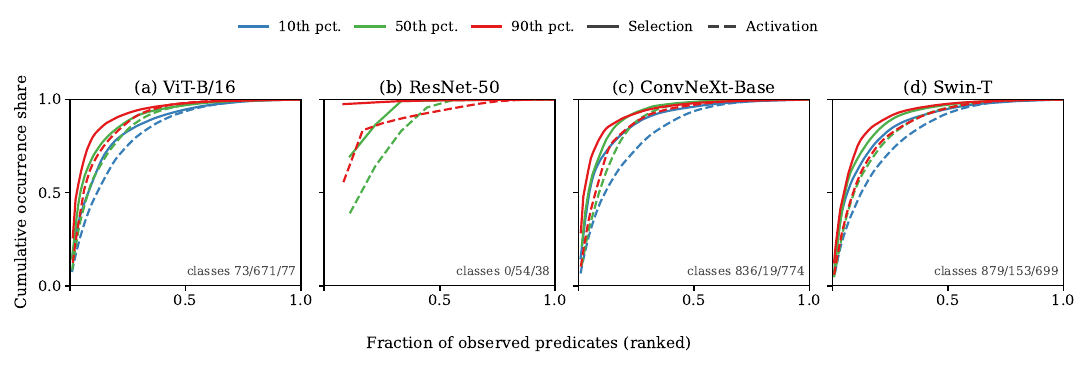}
    \caption{\textbf{Within-class concentration of predicate activations and selections on source images.} Predicates are ranked separately by activation and selection frequency, and curves show the cumulative share of occurrences. Colors identify classes nearest the 10th, 50th, and 90th percentiles of selection concentration, measured by $1-N_{80}/|\mathcal V_c|$. Solid lines show selection and dashed lines activation; both axes are normalized to $[0,1]$.}
    \vspace{-10pt}
    \label{fig:section42_concentration_appendix}
\end{figure}

\paragraph{Within-class concentration.}
For ViT, ResNet, ConvNeXt, and Swin, the median source vocabulary contains 95, 7, 81, and 133 predicates, respectively. However, only a small subset accounts for most selections: the median number needed to cover 80\% of within-class selected occurrences, $N_{80}$, is 16, 1.5, 11, and 23. The median fractions of the vocabulary needed to reach this coverage are 0.165, 0.222, 0.136, and 0.177 for selection, compared with 0.225, 0.364, 0.211, and 0.257 for activation. Thus, selections concentrate on fewer predicates than activations across all four models. Figure~\ref{fig:section42_concentration_appendix} illustrates classes near the 10th, 50th, and 90th percentiles of selection concentration. In the smaller held-out class groups, the relative concentration of selection and activation varies, so this ordering is clearest in the source results.

\paragraph{Predicates on held-out images.}
We assess how well the source predicates cover features selected on unseen images. A selected occurrence is \emph{eligible} when the source vocabulary contains a predicate for its predicted class and signed feature. The eligible fractions of all held-out selected occurrences are 98.566\%, 99.512\%, 98.920\%, and 99.468\% for ViT, ResNet, ConvNeXt, and Swin, respectively. Among these eligible occurrences, 97.357\%, 98.927\%, 98.359\%, and 98.428\% also satisfy their fixed source thresholds; we call this \emph{threshold retention}. Source retention is 100\% because the thresholds are fitted from selected source activations. The high held-out coverage and retention show that most selected features on unseen images can be represented by the source predicates without refitting, supporting the feature-selection and predicate-construction procedure in Sections~\ref{sec:pathways}--\ref{sec:predicates}.

\begin{figure}[!ht]
    \centering
    \includegraphics[width=\linewidth]{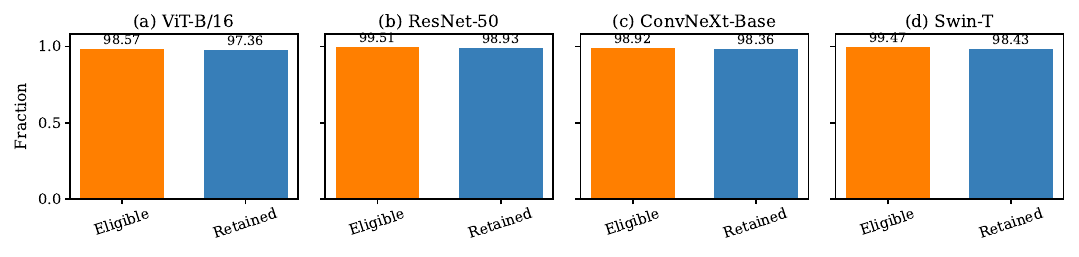}
    \caption{\textbf{Source predicate coverage and threshold retention on held-out images.} Eligible shows the percentage of all selected occurrences with a matching class-specific source predicate. Retained shows the percentage of eligible occurrences that satisfy the corresponding fixed threshold. Values are printed above the bars, showing how well the source predicates cover selected features and retain activation on unseen images.}
    \label{fig:section42_retention_appendix}
\end{figure}
\FloatBarrier

\section{Human evaluation}
\label{appdx: human}

\subsection{Experiment setup}
\label{appdx: human_eval}

We describe the evaluation setup in detail as follows.

\paragraph{Explanation stimuli.} The \textsc{VisionLogic} explanations shown to participants were generated using the bounding-box grounding procedure in Section~\ref{sec:grounding}, with the settings specified in Appendix~\ref{sec:grounding_details}. We used the accepted bounding boxes to show the grounded concepts, without the optional foreground-segmentation overlays. The \textsc{ACE} and \textsc{CRAFT} explanations were also presented using bounding boxes. Using a common region-display format helps control presentation differences when comparing concept usefulness.

\paragraph{Participants.} We recruited participants from Prolific \citep{prolific}, a popular online platform for human evaluation of research projects. The users recruited have high qualifications as we requested a task acceptance rate greater than 98\%. Our questionnaire was approved by the Social Sciences, Humanities \& Education REB at University X, and all users provided informed consent before they started the experiment. The questionnaire was designed to be completed in 8-10 minutes and users who passed the screening received US \$ 2.00 upon completion. 

\paragraph{Statistics.} For the Husky vs. Wolf scenario, $n =161$ participants passed screening and filtering, respectively $n=31, 25, 40, 35, 30$ for control, baseline, \textsc{ACE}, \textsc{CRAFT} and \textsc{VisionLogic}.

For the Otter vs. Beaver scenario\footnote{This task replaces the legacy ``Leaves'' dataset from \citet{DBLP:conf/nips/ColinFCS22} that is no longer available.}, $n = 114$ participants passed screening and filtering, respectively $n=22, 22, 27, 25, 18$ for control, baseline, \textsc{ACE}, \textsc{CRAFT}, and \textsc{VisionLogic}.

For the Kit Fox vs. Red Fox scenario, $n = 190$ participants passed screening and filtering, respectively $n=35, 35, 45, 40, 35$ for control, baseline, \textsc{ACE}, \textsc{CRAFT}, and \textsc{VisionLogic}.

\paragraph{Study design.} We followed the experimental design proposed by \citet{DBLP:conf/nips/ColinFCS22}, which quantitatively assesses to what degree a concept-based explanation method can help a human observer understand the behavior of an AI model.  Each participant is only tested on a single condition (control, baseline, \textsc{ACE}, \textsc{CRAFT}, or \textsc{VisionLogic}) to avoid possible experimental confounds.



\begin{figure}[!h]
    \centering
    \includegraphics[width=\linewidth]{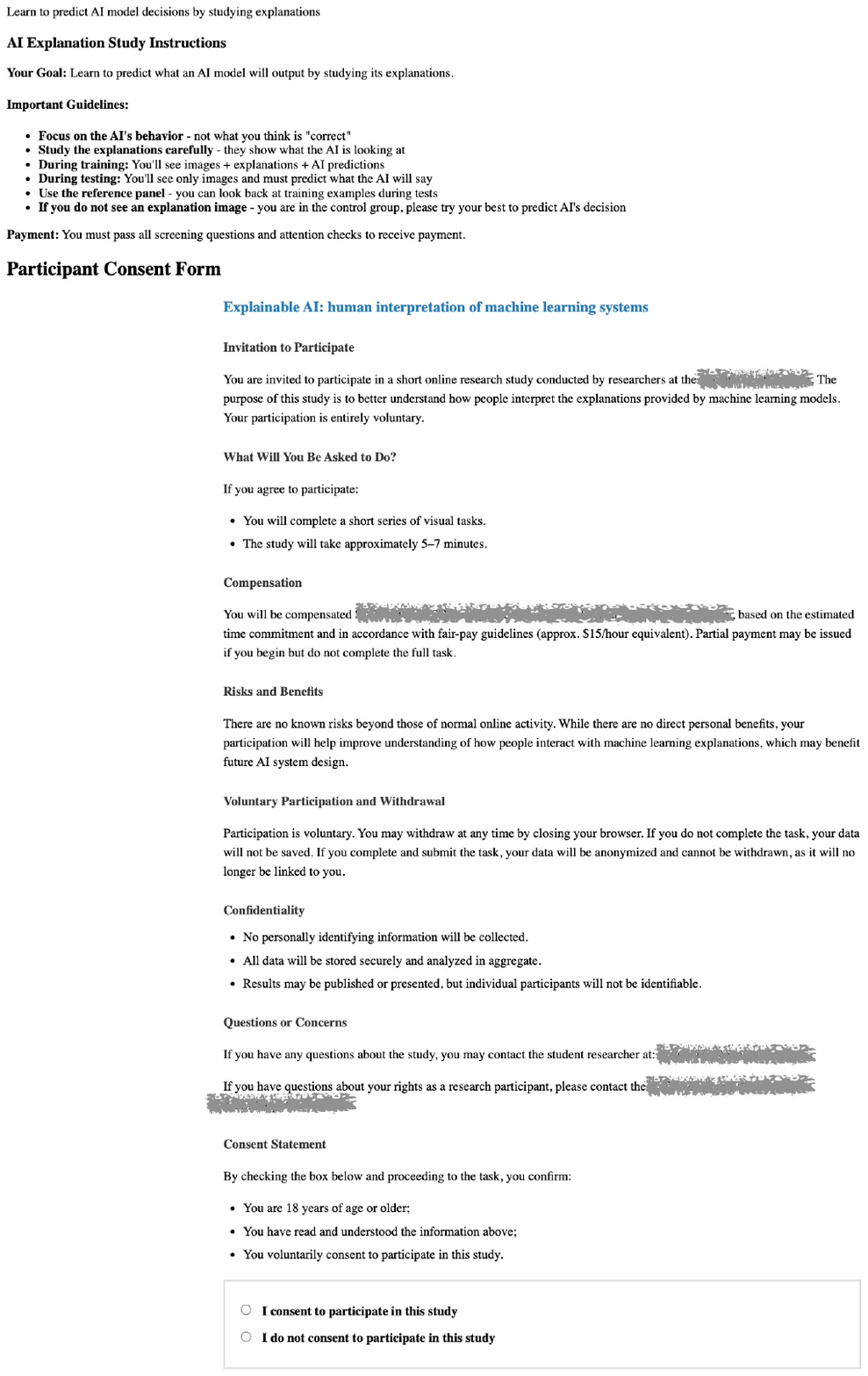}
    \caption{The online questionnaire begins with a consent form.}
    \label{fig:consent}
\end{figure}

The screening, training, and testing phases follow the procedure described in \citet{DBLP:conf/nips/ColinFCS22}. After granting consent (shown in Figure~\ref{fig:consent}),  participants are presented with detailed instructions and the study goal: learning to predict which class \textit{an AI model will predict} for a given image. They start with a practice session that shows simple images along with their explanations and model prediction, followed by some unseen images without explanations, where they are expected to answer the model's prediction correctly. Participants who failed this session were not allowed to proceed with the study. They are then tested on their understanding of the study goal by a short quiz (exactly formulated as in \citet{DBLP:conf/nips/ColinFCS22}). Again, participants who failed this session were not allowed to proceed. 

\begin{figure}[!h]
    \centering
    \includegraphics[width=0.9\linewidth]{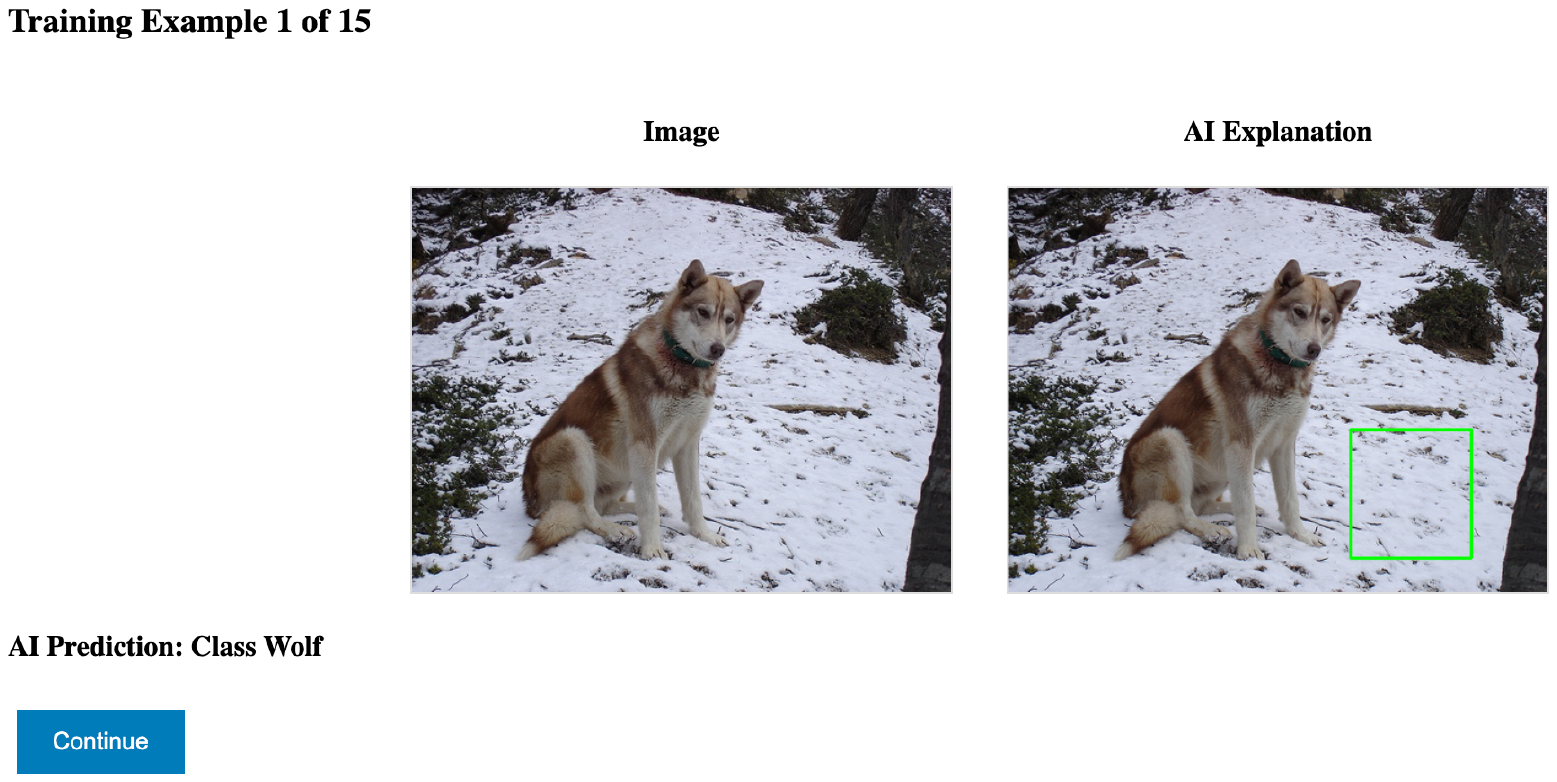}
    \caption{The online questionnaire displaying an example of the training session with the original image, the explanation, and the model prediction.}
    \label{fig:exp_eg}
\end{figure}

\begin{figure}[!h]
    \centering
    \includegraphics[width=0.8\linewidth]{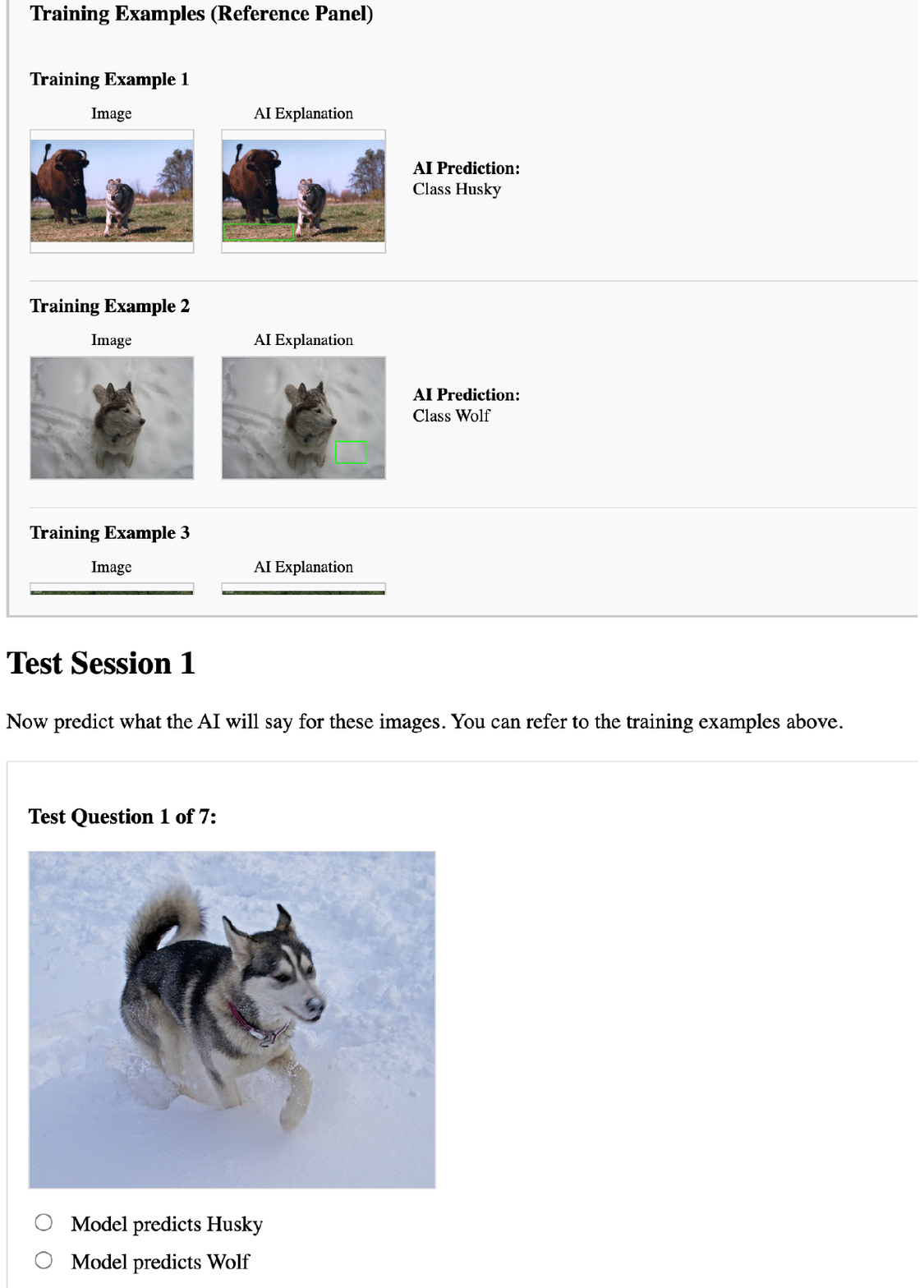}
    \caption{The online questionnaire displaying the testing session with a reservoir containing all examples during the training phase.}
    \label{fig:ref_panel}
\end{figure}

After the above two screening phases, participants went through 3 sessions of training phases (5, 10, and 15 images, respectively, with explanations and model decisions) followed by a testing phase (7 new images without explanation). The answers for the testing phases are collected to compute the main result. Figure~\ref{fig:exp_eg} shows an example of the training phase. As described in \citet{DBLP:conf/nips/ColinFCS22}, we implemented a reservoir (Figure~\ref{fig:ref_panel}) for training images as a reference point for participants during the testing phase. The last test image of each session is an image from the reservoir, and participants who incorrectly answered the last test question were filtered out.

\subsection{Statistical test results for significance}
\label{appdx:human_eval_stats}

\begin{figure}[!h]
    \centering
    \includegraphics[width=0.8\linewidth]{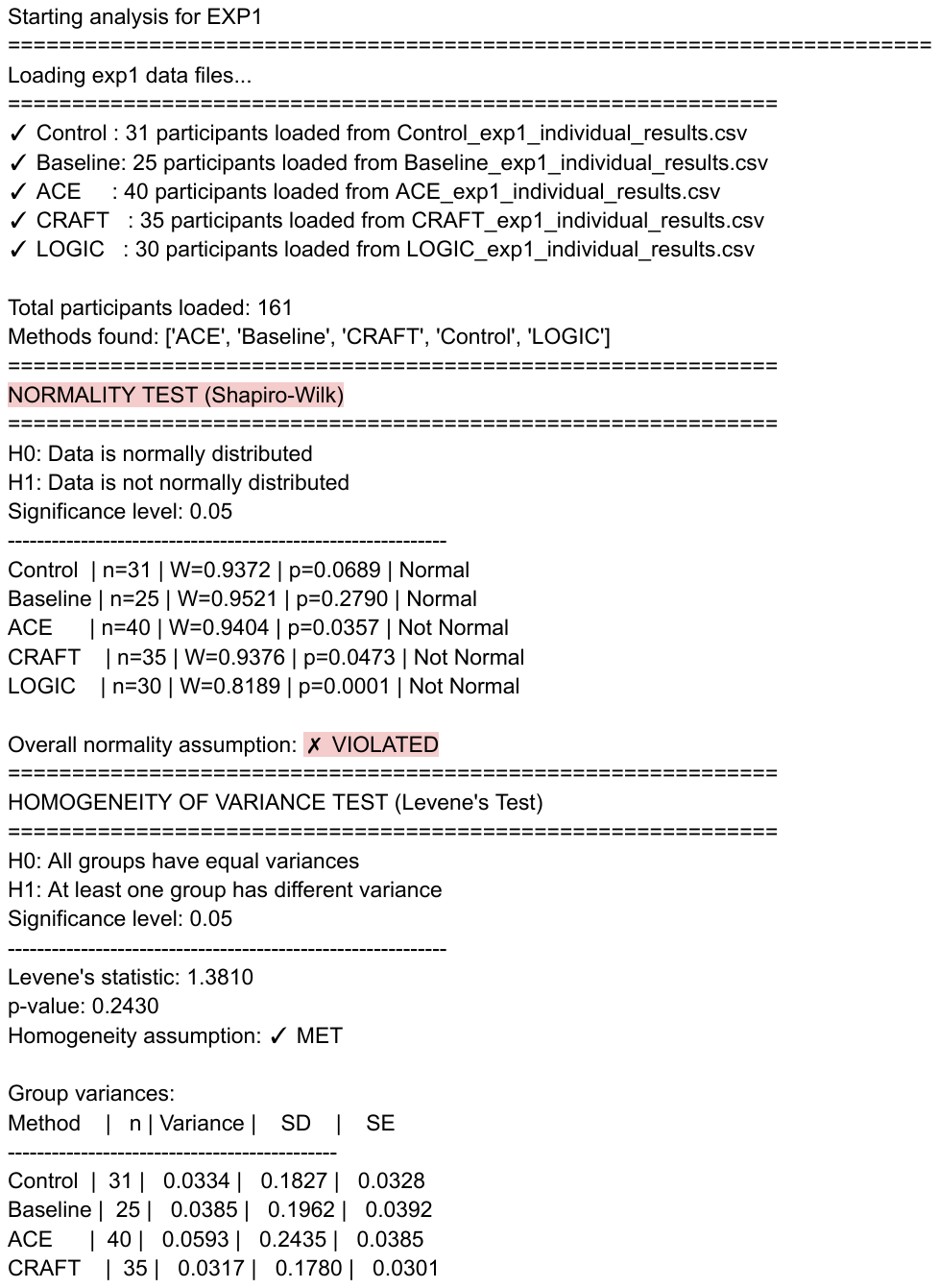}
    \caption{Complete statistical test results for scenario 1, page 1.}
    \label{fig:stats_test_restult_exp1_p1}
\end{figure}

\begin{figure}[!h]
    \centering
    \includegraphics[width=0.8\linewidth]{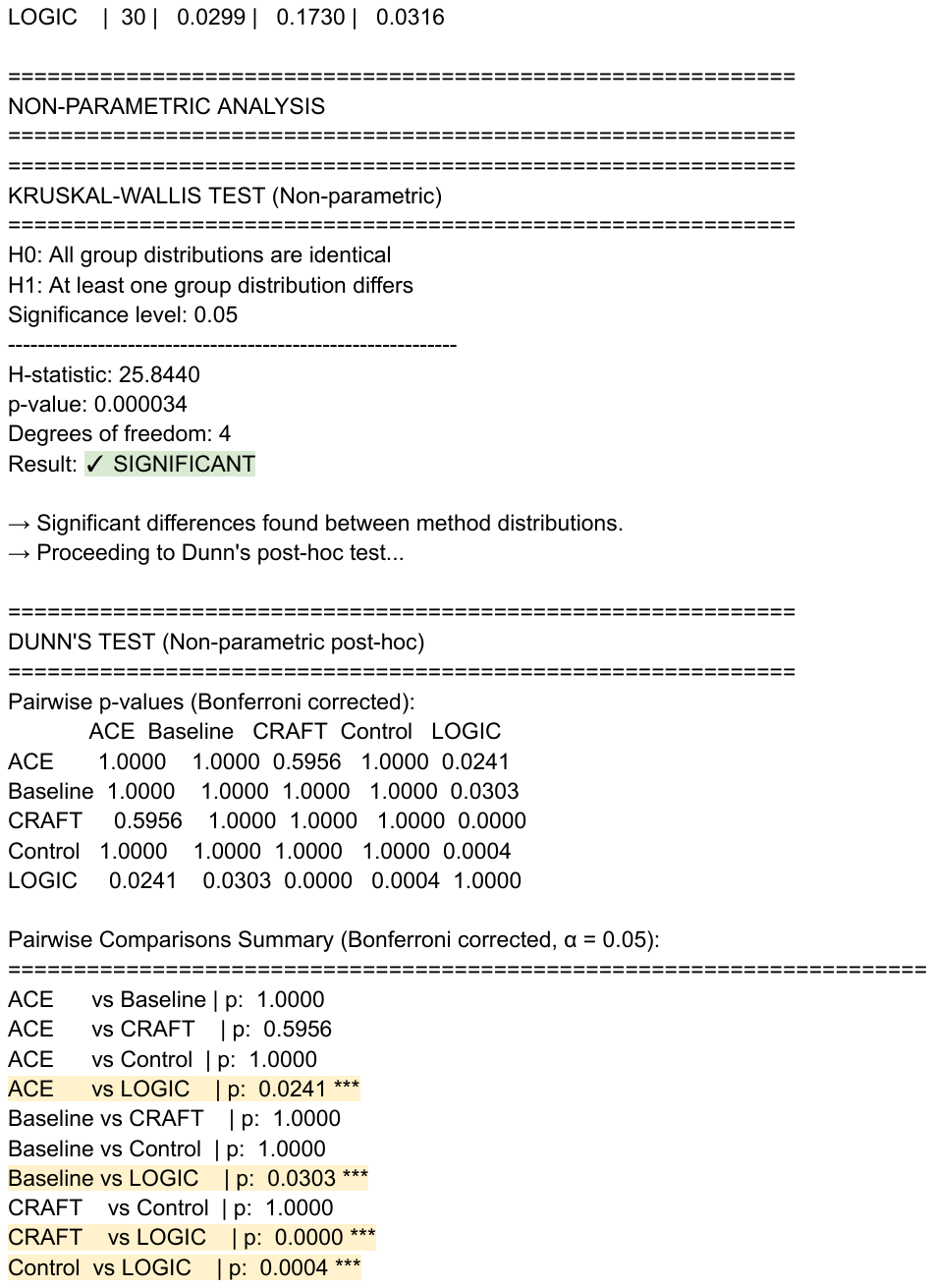}
    \caption{Complete statistical test results for scenario 1, page 2.}
    \label{fig:stats_test_restult_exp1_p2}
\end{figure}

\begin{figure}[!h]
    \centering
    \includegraphics[width=0.8\linewidth]{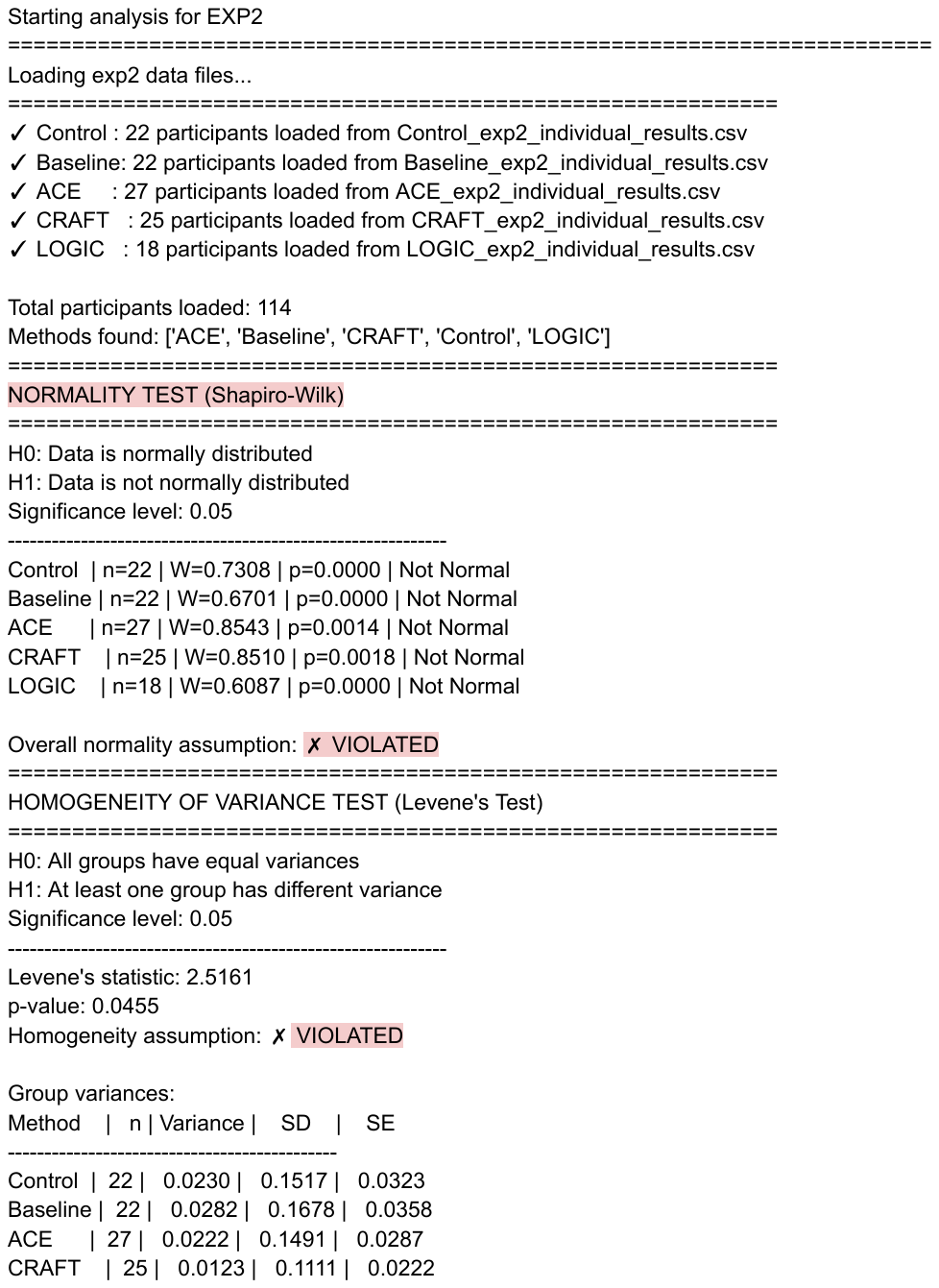}
    \caption{Complete statistical test results for scenario 2, page 1.}
    \label{fig:stats_test_restult_exp2_p1}
\end{figure}

\begin{figure}[!h]
    \centering
    \includegraphics[width=0.8\linewidth]{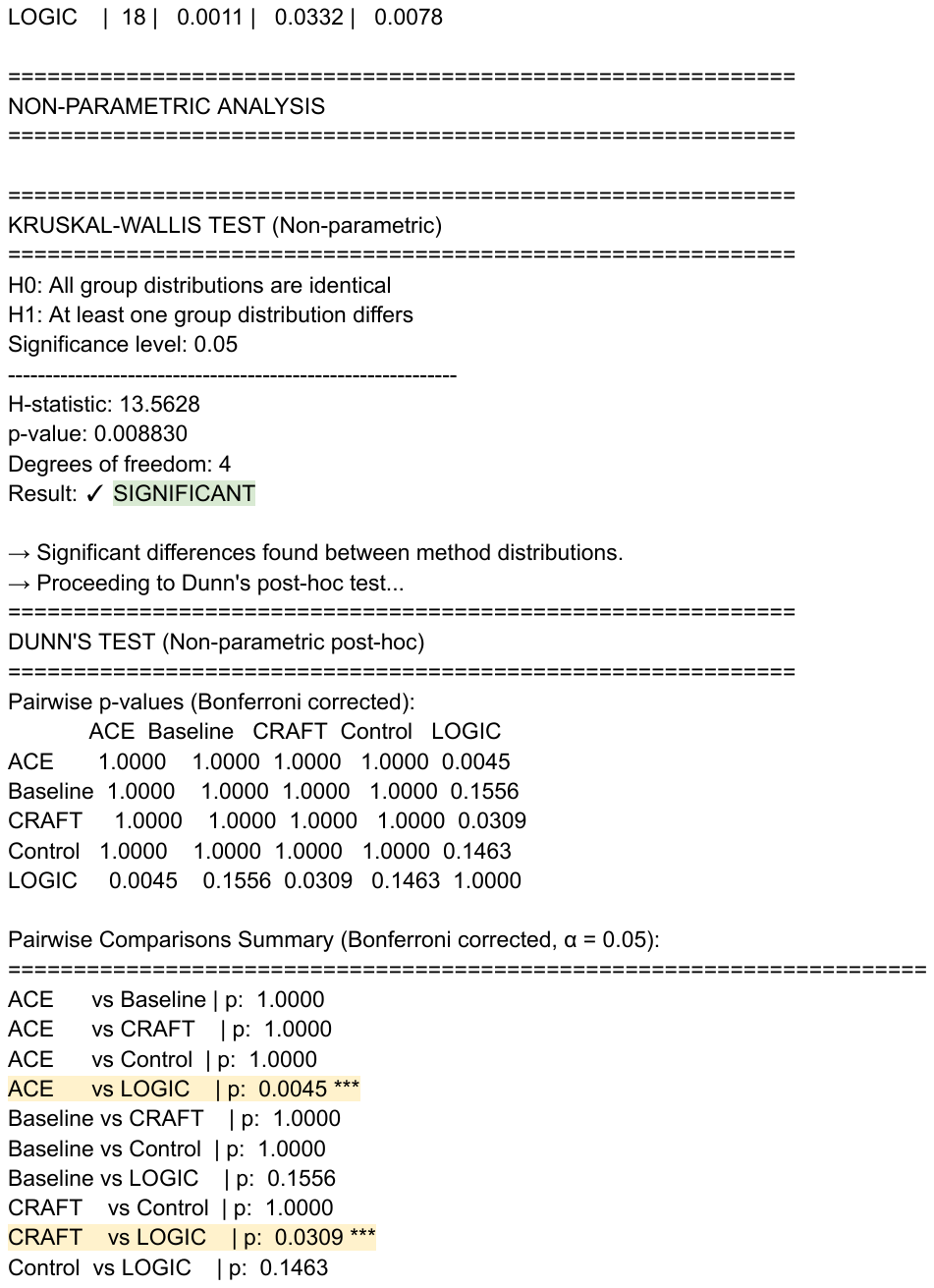}
    \caption{Complete statistical test results for scenario 2, page 2.}
    \label{fig:stats_test_restult_exp2_p2}
\end{figure}

\begin{figure}[!h]
    \centering
    \includegraphics[width=0.8\linewidth]{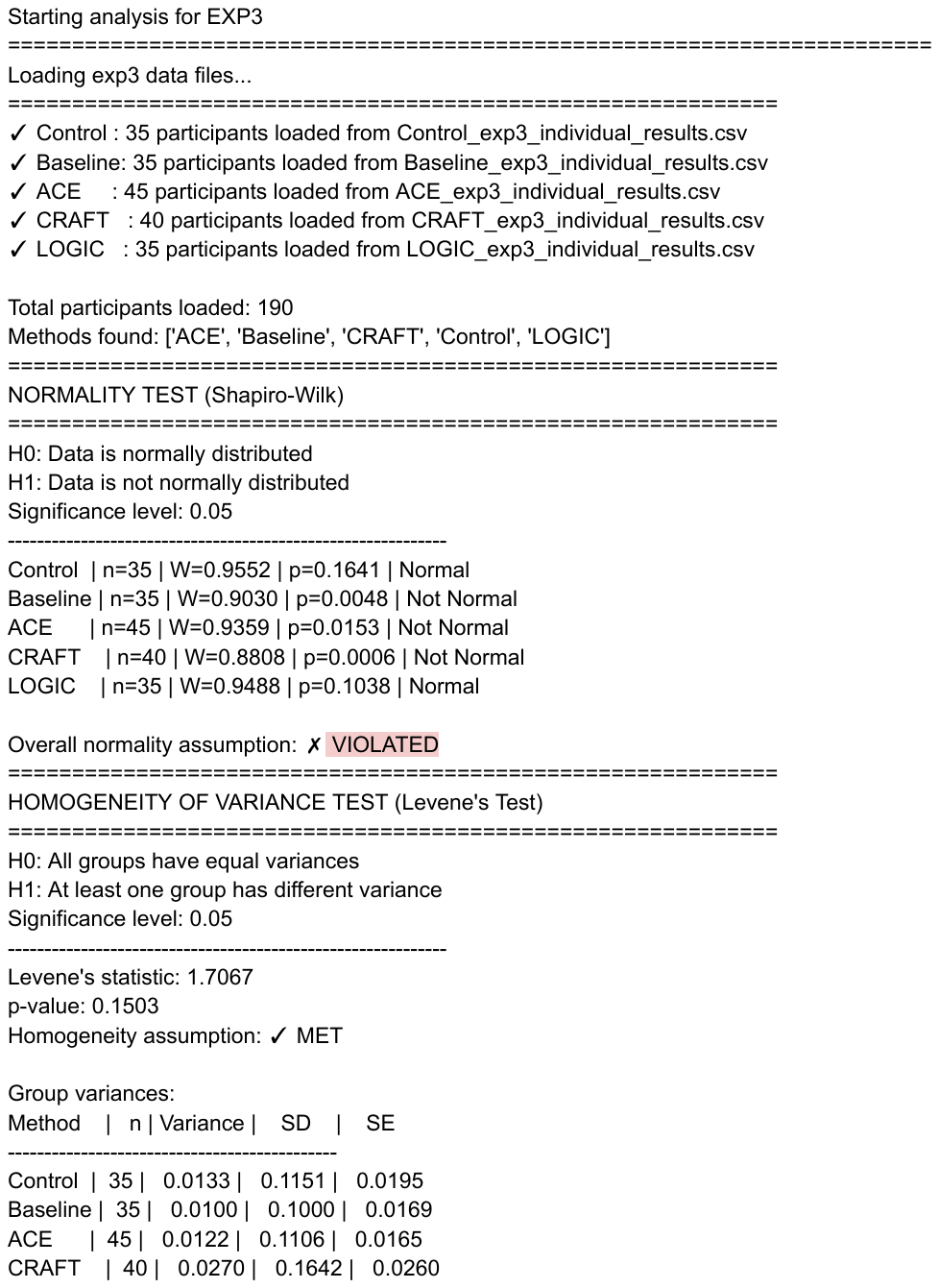}
    \caption{Complete statistical test results for scenario 3, page 1.}
    \label{fig:stats_test_result_exp3_p1}
\end{figure}

\begin{figure}[!h]
    \centering
    \includegraphics[width=0.8\linewidth]{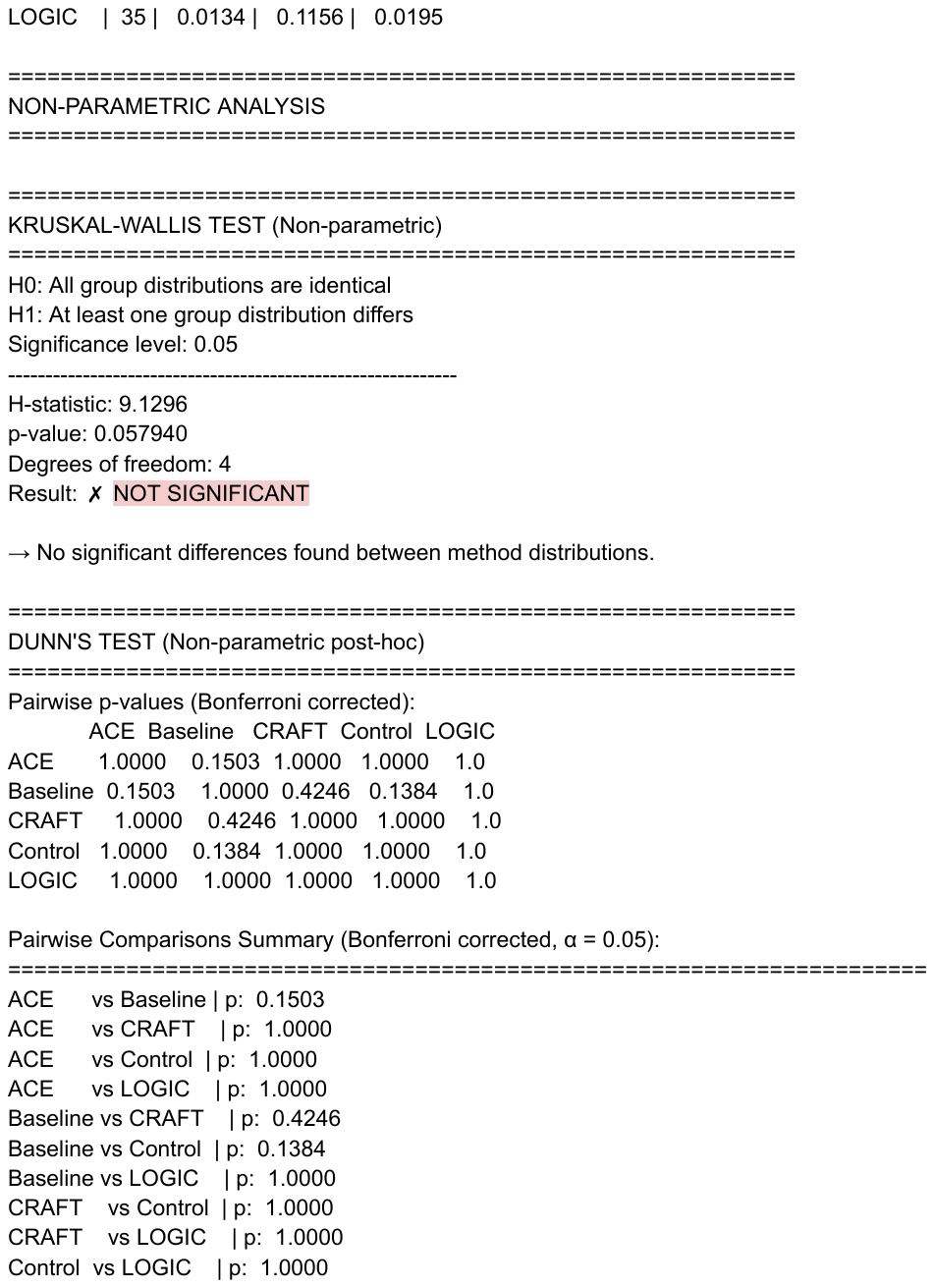}
    \caption{Complete statistical test results for scenario 3, page 2. The overall test does not reject at the 0.05 level.}
    \label{fig:stats_test_result_exp3_p2}
\end{figure}

We provide the statistical test results in Figures~\ref{fig:stats_test_restult_exp1_p1}, \ref{fig:stats_test_restult_exp1_p2}, \ref{fig:stats_test_restult_exp2_p1}, \ref{fig:stats_test_restult_exp2_p2}, \ref{fig:stats_test_result_exp3_p1}, and~\ref{fig:stats_test_result_exp3_p2}.


\end{document}